\documentclass[10pt,onecolumn,letterpaper]{article}
\usepackage{cvpr}
\usepackage{algpseudocode}%
\usepackage{enumitem} 
\usepackage{listings}%
\usepackage{url}
\usepackage{array}
\usepackage{adjustbox}%
\usepackage{times}
\usepackage{epsfig}
\usepackage{graphicx}
\usepackage{amsthm}
\usepackage{textcomp}%
\usepackage{cite}
\usepackage{stfloats}
\usepackage{verbatim}
\usepackage{booktabs}%
\usepackage{tcolorbox} %
\usepackage[T1]{fontenc}
\tcbuselibrary{breakable}
\usepackage{multirow}%
\usepackage{amsmath,amssymb,amsfonts}%
\usepackage{algorithmicx}%
\usepackage{xcolor}%
\usepackage[utf8]{inputenc}
\usepackage{newunicodechar}
\newunicodechar{−}{\textminus}
\usepackage{amssymb}
\tcbuselibrary{raster}
\usepackage[breaklinks=true,bookmarks=false]{hyperref}

\numberwithin{equation}{subsection}
\usepackage{datetime}
\usepackage[a4paper,margin=1in,headheight=15pt,headsep=24pt]{geometry}

\usepackage{fancyhdr}

\cvprfinalcopy 
\def\cvprPaperID{****} 
\allowdisplaybreaks

\begin{document}
\pagestyle{fancy}
\lhead{}

\title{Re-Initialization Token Learning for Tool-Augmented Large Language Models}
\author{
\textbf{Chenghao Li}$^{1,2}$\quad
\textbf{Liu Liu}$^{1,2*}$\quad
\textbf{Baosheng Yu}$^{3}$\quad
\textbf{Jiayan Qiu}$^{4}$\quad
\textbf{Yibing Zhan}$^{5}$\\
[0.5em]
$^1$School of Artificial Intelligence, Beihang University\\
$^2$Hangzhou International Innovation Institute, Beihang University\\
$^3$Nanyang Technological University \quad
$^4$University of Leicester\\
$^5$Yunnan United Vision Technology
}
\maketitle
\begingroup
\renewcommand\thefootnote{*}
\footnotetext{Corresponding author: \texttt{liuliubh@buaa.edu.cn}

Our code are publicly available at \url{https://github.com/lichenghaobuaa/TokenLearning}
}
\endgroup
\thispagestyle{fancy}
\begin{abstract}
Large language models have demonstrated exceptional performance, yet struggle with complex tasks such as numerical reasoning, plan generation. Integrating external tools, such as calculators and databases, into large language models (LLMs) is crucial for enhancing problem-solving capabilities. 
Current methods assign a unique token to each tool, enabling LLMs to call tools through token prediction—similar to word generation. However, this approach fails to account for the relationship between tool and word tokens, limiting adaptability within pre-trained LLMs.  
To address this issue, we propose a novel token learning method that aligns tool tokens with the existing word embedding space from the perspective of initialization, thereby enhancing model performance.
We begin by constructing prior token embeddings for each tool based on the tool’s name or description, which are used to initialize and regularize the learnable tool token embeddings. This ensures the learned embeddings are well-aligned with the word token space, improving tool call accuracy. 
We evaluate the method on tasks such as numerical reasoning, knowledge-based question answering, and embodied plan generation using GSM8K-XL, FuncQA, KAMEL, and VirtualHome datasets. The results demonstrate clear improvements over recent baselines, including CoT, REACT, ICL, and ToolkenGPT, indicating that our approach effectively augments LLMs with tools through relevant tokens across diverse domains.

\end{abstract}

\section{Introduction}\label{sec1}

Large language models (LLMs) \cite{brown2020languagemodelsfewshotlearners}\cite{chen2024agentflandesigningdatamethods} have emerged as powerful tools across a wide range of applications, from content generation to customer service \cite{bommarito2022gpttakesbarexam,borgeaud2022improvinglanguagemodelsretrieving,li2023modelscopeagentbuildingcustomizableagent}. As the technology behind these models advances, there is growing interest in their ability to integrate with external tools \cite{parisi2022talmtoolaugmentedlanguage,ye2024tooleyesfinegrainedevaluationtool,chowdhery2022palmscalinglanguagemodeling,9782500}, such as computational aids and data repositories \cite{thoppilan2022lamdalanguagemodelsdialog,schick2023toolformerlanguagemodelsteach}. The ability of LLMs to leverage a broad spectrum of tools not only demonstrates their cognitive potential but also helps address inherent limitations\cite{tang2023toolalpacageneralizedtoollearning}, such as staying current with global knowledge\cite{qin2023webcpminteractivewebsearch}, reducing inaccurate information generation \cite{roller2020recipesbuildingopendomainchatbot}\cite{touvron2023llamaopenefficientfoundation}, and performing complex symbolic tasks. 
The constant emergence of new tools, including advanced software frameworks and domain-specific utilities \cite{liang2023taskmatrixaicompletingtasksconnecting}, adds complexity to tool acquisition for LLMs. To address this,  two primary approaches have been proposed for integrating tools into LLMs \cite{mialon2023augmentedlanguagemodelssurvey}. The first approach fine-tunes LLMs to learn specific tools \cite{parisi2022talmtoolaugmentedlanguage}. While effective in some cases, this method is computationally expensive and struggles to adapt to new tools. The second approach, in-context learning, enables LLMs to handle new tools and has been successfully applied in various scenarios. However, it remains limited by the context length and performs poorly when mastering new tools with only a few examples\cite{rozière2024codellamaopenfoundation}\cite{zeng2022socraticmodelscomposingzeroshot}.

\begin{figure}[!t]
    \centering
    \includegraphics[width=0.85\linewidth]{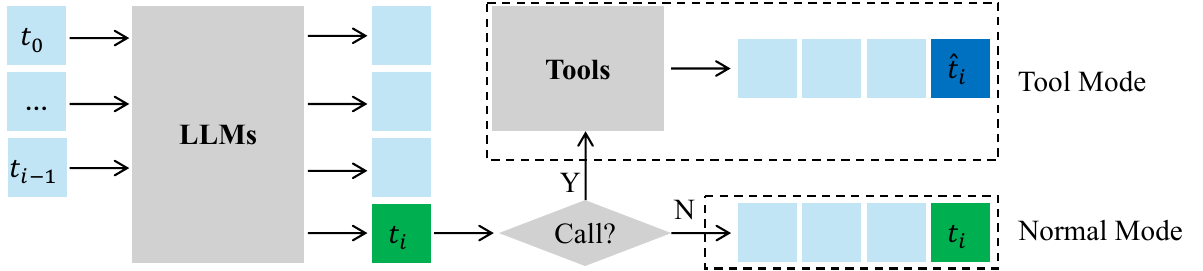}
    \caption{An illustration of tool-augmented large language models. After inputting the command text $<t_0,...,t_{i-1}>$ segment to the LLMs, the LLM appended $t_i$ to the output segment. This serves as an indicator to determine whether tool invocation is required. If tool usage is unnecessary, the system switches to Normal Mode and directly outputs the result. If tool invocation is required, the system transitions to Tool Mode and subsequently outputs the processed results from the tool.}
    \label{fig:tool-llm}
\end{figure}

Recently, the ToolkenGPT method~\cite{hao2024toolkengptaugmentingfrozenlanguage} was introduced to enhance LLMs by embedding multiple tools, enabling seamless integration via learned tool tokens. Figure~\ref{fig:tool-llm} illustrates this tool-augmented LLM framework. By introducing additional tool tokens, the system supports two modes for next-token prediction: 1) If the predicted token is a word token, the system operates in the standard mode~\cite{hao2024toolkengptaugmentingfrozenlanguage}; 2) If the predicted token corresponds to a tool, the system switches to tool mode and generates the tool'output as the next token~\cite{hao2024toolkengptaugmentingfrozenlanguage}. Thus, the effectiveness of the learned tool tokens is critical for the success of this mode switch. Current token learning approaches typically learn token embeddings from scratch before integrating them with the vocabulary of tokens~\cite{hao2024toolkengptaugmentingfrozenlanguage}. However, such  approaches overlook the semantic relationship between tool and word token embeddings \cite{li2023apibankcomprehensivebenchmarktoolaugmented}, which limits its adaptability within pre-trained LLMs \cite{huang2022innermonologueembodiedreasoning}.

To address this limitation, we propose a novel token learning approach that jointly optimizes tool token embeddings for next-token prediction while ensuring their alignment with the word embedding space  through a re-initialization perspective.
Following ToolkenGPT~\cite{hao2024toolkengptaugmentingfrozenlanguage}, we construct training sequences that integrate both word and tool tokens, where tool-related tokens replace specific subsequences.   To enhance consistency with the word embedding space, we align the learned tool token embeddings with prior tool token embeddings derived from word tokens. Specifically, these prior embeddings are constructed as follows. For each tool, we begin by extracting one or more word tokens from its name or description. We then calculate the tool’s prior embedding by averaging the embeddings of these extracted word tokens. This prior embedding serves to regularize the optimization of the learnable tool token embedding, ensuring alignment with the prior embedding, i.e., the word embedding space. Notably, the prior embeddings also serves as initialization for the learnable tool token embeddings, which helps accelerate convergence. As a result, the regularized token learning approach facilitates the learning of effective tool token embeddings that align with the existing word embeddings used by LLMs.

To evaluate our proposed token learning approach for tool-augmented LLMs, we conducted comprehensive experiments across three representative tasks: mathematical problem solving, knowledge-based question answering, and embodied plan generation. In each of these tasks, external tools play a crucial role by significantly enhancing the reasoning capabilities of LLMs. Our results demonstrate that the proposed tool token learning approach significantly improves LLMs’ tool selection accuracy for complex problems, especially those involving requiring numerical calculations. 
Furthermore, the results highlight the importance of maintaining consistency between additional token embeddings and the original vocabulary when augmenting pre-trained LLMs. In other words, the information contained in the original vocabulary can substantially enhance the model’s ability to master and effectively use new tools.
The main contributions of our proposed framework are as follows:
\begin{itemize}
    \item We propose a novel token learning approach for tool-augmented LLMs, which significantly enhances the accuracy of LLMs in selecting appropriate tools for complex tasks, particularly in scenarios requiring numerical calculations. 
    \item We introduce a pooling-based token embedding method to connect tool tokens with the LLM vocabulary, especially in complex scenarios. A regularization term is added to the loss function to ensure that the learned embeddings remain close to the prior embeddings.
    \item Empirical evaluations on three representative tasks: \textit{mathematical problem solving}, \textit{knowledge-based question answering}, and \textit{embodied plan generation} across LLaMA-2 models (7B, 13B, and 70B).
    In the tasks of mathematical problem solving, our method has improved the accuracy by approximately 3\% compared to the latest method, ToolkenGPT. In the other two tasks, our method further improves the accuracy of the model in tool invocation, especially when the number of tools is large and the success rate of generated plan is low.
\end{itemize}
\section{Related Work}\label{sec2}

\subsection{Tool Tokenization Paradigms: The ToolkenGPT Approach}
ToolkenGPT \cite{hao2024toolkengptaugmentingfrozenlanguage} represents a significant advancement in tool integration for large language models (LLMs), introducing an innovative tokenization paradigm that addresses key limitations of previous approaches. By formulating tools as special tokens called "toolkens," this method enables seamless integration of external tools into the standard text generation process. Each toolken functions similarly to a word token but is associated with an embedding vector that encapsulates the tool's functionality.
The operational mechanism of ToolkenGPT\cite{hao2024toolkengptaugmentingfrozenlanguage} involves several sophisticated steps: when the model predicts a toolken during generation, it enters a specialized mode where it generates appropriate input arguments for the corresponding tool. This transition is managed through carefully designed prompting strategies that maintain the model's contextual understanding while adapting to tool-specific requirements. After receiving the tool's output, the system reintegrates this information into the ongoing generation process, creating a smooth interaction between language modeling and tool execution.

This approach demonstrates particular strength in three key application areas: numerical reasoning tasks where precise calculations are required, knowledge-based question answering that benefits from external data sources, and embodied plan generation that requires interaction with simulated environments. The tokenized tool representation allows ToolkenGPT to outperform traditional methods like Chain-of-Thought \cite{wei2023chainofthoughtpromptingelicitsreasoning} and ReAct \cite{yao2023reactsynergizingreasoningacting} by eliminating the need for verbose intermediate reasoning steps while maintaining precise tool control.

\subsection{Evolution of Fine-tuning Based Tool Integration}
The historical development of tool integration in LLMs reveals a clear progression from specialized, fine-tuned systems to more flexible approaches. Early efforts in this domain primarily relied on model fine-tuning to achieve tool competency, focusing on enabling LLMs to work with a constrained set of tools within well-defined domains. Retrieval mechanisms emerged as one of the most impactful early tools, with systems like REALM \cite{guu2020realmretrievalaugmentedlanguagemodel}, RAG \cite{lewis2021retrievalaugmentedgenerationknowledgeintensivenlp}, and RETRO \cite{borgeaud2022improvinglanguagemodelsretrieving} demonstrating how external knowledge sources\cite{10706809} \cite{Yu_2022} could significantly enhance model performance on knowledge-intensive tasks.
The WebGPT \cite{nakano2022webgptbrowserassistedquestionansweringhuman} system marked an important milestone by showing how human-like web search behaviors could be effectively incorporated into LLMs through fine-tuning. This work paved the way for broader tool integration efforts, with subsequent research expanding the range of incorporated tools to include question-answering systems, computational tools like calculators, language translation services, and various other utilities. Notable contributions in this expansion include TALM \cite{parisi2022talmtoolaugmentedlanguage}, which systematically explored tool augmentation across multiple domains, and Toolformer \cite{schick2023toolformerlanguagemodelsteach}, which introduced self-supervised learning for tool use.

Despite these advances, the fine-tuning paradigm presents fundamental limitations that become increasingly apparent as the field progresses. The computational resources required for effective fine-tuning grow substantially with model size, creating significant barriers to widespread adoption. Furthermore, fine-tuned models exhibit limited flexibility when facing new tools or updated versions of existing tools, often requiring complete retraining to maintain functionality.

\subsection{In-Context Learning for Tool Usage}
The exploration of in-context learning for tool usage represents a paradigm shift from the fine-tuning approaches discussed earlier. This methodology capitalizes on LLMs' remarkable ability to learn from contextual examples, eliminating the need for weight updates while maintaining flexibility. The approach works by embedding tool descriptions and usage demonstrations directly within the prompt structure \cite{mialon2023augmentedlanguagemodelssurvey}\cite{qin2024toollearningfoundationmodels}, allowing models to adapt their behavior dynamically based on the provided examples.
Practical implementations of this approach, such as those seen in ChatGPT plugins, demonstrate its potential for real-world applications. A typical usage scenario might involve showing the model multiple examples of calculator tool usage\cite{schick2023toolformerlanguagemodelsteach}, including the precise format for input expressions and output interpretations. While effective for simple tools and common use cases\cite{qin2024toollearningfoundationmodels}, this method encounters significant challenges when dealing with more complex scenarios. The finite context window of current LLMs imposes strict limits on the number and complexity of tools that can be effectively demonstrated, while the few-shot learning paradigm often proves insufficient for reliable tool mastery.
REACT \cite{yao2023reactsynergizingreasoningacting}\cite{10387715}\cite{10506571} offers a complementary approach that structures tool interaction through predefined action spaces. In knowledge-intensive applications, REACT typically employs a set of fundamental actions including search, lookup, and finish operations, often implemented through standardized APIs like Wikipedia's interface. The system's effectiveness is particularly evident in tasks like HotPotQA \cite{yang2018hotpotqadatasetdiverseexplainable}, where the model's reasoning process directly informs its tool usage strategy.

However, REACT's reliance on predefined action spaces creates its own set of constraints. Complex, multi-step tasks often exceed the system's capacity due to context window limitations \cite{zhong2023memorybankenhancinglargelanguage}\cite{10937907}, while the need for careful action space design introduces additional implementation complexity. These limitations highlight the ongoing challenges in developing truly flexible and scalable tool integration methods for modern LLMs.

\begin{figure*}[t]
\centering
\includegraphics[width=1\linewidth]{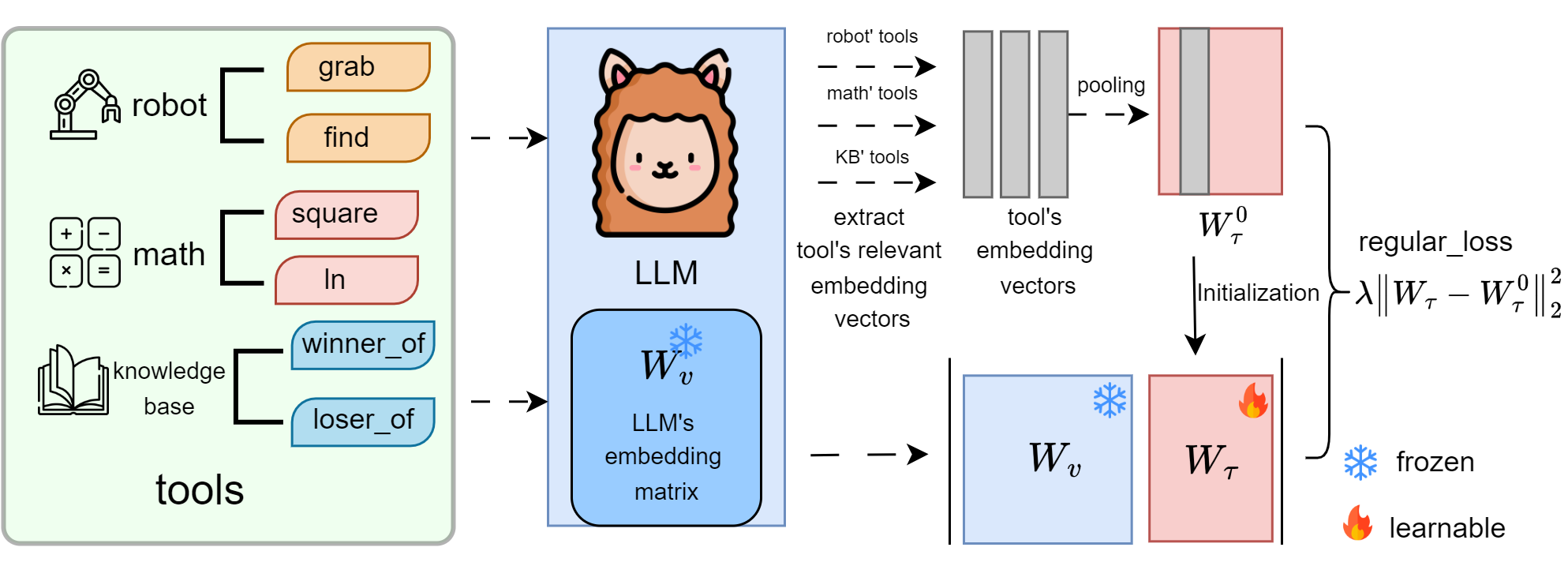}
\caption{The TokenLearning framework operates through the following methodological pipeline: First, we extract tool-related embedding vectors from the pretrained language model's vocabulary matrix \(\textbf{W}_v\). These extracted embeddings then undergo a pooling operation to aggregate their feature representations. Subsequently, we concatenate the processed embedding vectors corresponding to each individual tool to construct the initial matrix \(\textbf{W}_{\tau}^0\). This constructed matrix serves dual purposes: (1) as the initialization value for the learnable tool embedding matrix \(\textbf{W}_{\tau}\), and (2) as a regularization constraint during optimization. Through this approach, the final optimized "toolken" matrix \(\textbf{W}_{\tau}\) appended to the large language model exhibits enhanced directional properties, enabling more precise tool invocation capabilities. }
\label{fig:OverviewofTokenlearnframework}
\end{figure*}

\section{Method}\label{sec3}

This section introduces a regularized token learning framework for tool-augmented LLMs, aiming to align tool token embeddings with word embedding spaces and improve tool invocation accuracy. First, prior embeddings are constructed from tool names to initialize learnable tokens. Second, pooling operations (e.g., average/max pooling) aggregate word token features for embedding alignment. Finally, a regularization term constrains learned embeddings to match prior ones, enhancing training stability and generalization.

\subsection{Tool-Augmented LLMs}\label{subsec1}

LLMs model the probability of a sequence of word tokens \(s=(t_{1},t_{2},\cdots,t_{n})\) as \(P(s)=\sum_{i}^{n}P(t_{i}\vert t_{<i})\), where \(t_{i}\in \mathcal{V}\) and \(t_{<i}\) represents the partial sequence of tokens preceding the \(i\)-th token. The formula for predicting the distribution of the next word token is 
\begin{align}
    P(t_{i}\vert t_{<i})=\text{softmax}(\textbf{W}_{v}\cdot\textbf{h}_{i - 1}),
\end{align}
where \(\boldsymbol{h}_{i - 1}\in\mathbb{R}^{d}\) denotes the last hidden state of the current context, and \(\textbf{W}_{v}\in\mathbb{R}^{|\mathcal{V}|\times d}\) is the embedding matrix for word tokens, and $\mathcal{V}$ represents the complete set of word tokens in LLMs \cite{touvron2023llama2openfoundation}. The concept of tool tokens is also termed ``toolken” in ToolkenGPT \cite{hao2024toolkengptaugmentingfrozenlanguage}. Given a set of tools \(\mathcal{T}\), the next token prediction is then formulated as 
\begin{align}
    P(t_{i}\vert t_{<i})=\text{softmax}([\textbf{W}_{v};\textbf{W}_{\tau}]\cdot\boldsymbol{h}_{i - 1}),
\end{align}
where \(t_{i}\in \mathcal{V}\cup \mathcal{T}\) and \(\textbf{W}_{\tau}\in\mathbb{R}^{\vert \mathcal{T}\vert\times d}\) is the embedding matrix of tool tokens\cite{hao2024toolkengptaugmentingfrozenlanguage}.


When a tool invocation is triggered, the language model switches to the tool mode\cite{hao2024toolkengptaugmentingfrozenlanguage}, pauses the current text generation, and completes the parameter generation according to the context demonstrations of the tool in the prompt and the syntax \([\mathsf{tool}](\mathsf{arguments})\). After the tool is executed, the result is sent back to the text in the reasoning mode for further processing \cite{hao2024toolkengptaugmentingfrozenlanguage}.
Specifically, given a sentence, e.g., ``there are 100 dollars", it can be tokenized into a word token sequence \textit{s} = (``there”, ``are”, ``1”, ``0”, ``0”, ``dollars”). To indicate when to predict the toolkens, we need a parallel sequence mixed with word tokens and toolkens, i.e. \(s'\) = (``there”, ``are”,``[add]”,``[N/A]”, ``[N/A]”,``[N/A]”,``dollars”). The position of ( ``1”, ``0”, ``0”) in \textit{s} is where the returned tool's results fill in, and the corresponding first token in \(s'\) is the toolken for the tool call with the following tokens are filled with [N/A], indicating neglect in loss calculation. 
When ToolkenGPT learns toolken embeddings matrix \(\textbf{W}_{\tau}\), given a dataset \(\mathcal{D}\) composed of \((s,s')\), the training objective becomes
\begin{equation}
    \mathcal{L}(W_{\tau})=\sum_{(s,s')\in \mathcal{D}}\sum_{i = 1}^{N}-\log P(t_{i}'\vert t_{<i})\mathbb{I}_{t_{i}'\neq[\text{N/A}]},    
\end{equation}
where $P(t_{i}'\vert t_{<i})$ is calculated according to the above formula and $\mathbb{I}_{t_{i}'\neq[\text{N/A}]}$is used to ignore the [N/A] tokens during training~\cite{hao2024toolkengptaugmentingfrozenlanguage}.

\subsection{Prior Token Embeddings}

The main idea of regularized token learning is to explicitly link tool tokens to those in the vocabulary of LLMs that corresponds to tools. To achieve this, we first construct prior token embeddings \(\textbf{W}_{\tau}^0\) for each tool based on the tool’s name or description, which are then used to initialize and regularize the learnable tool token embeddings \(\textbf{W}_{\tau}\). For example, when solving mathematical problems, tools or operations such as \textit{add}, \textit{subtract}, \textit{multiply}, and \textit{divide} are used. Each of these tool names corresponds to a word token in the LLM’s vocabulary, and we directly extract their word embeddings to serve as the prior embeddings for these tools. In cases where a tool name maps to multiple word tokens, we apply global  pooling operations across the embeddings of these tokens to obtain a single prior embedding for the tool.

We can perform average pooling or max pooling on the multi-dimensional vectors to transform them into the embedding vectors corresponding to the relevant tools. Note that \(y\) is the output vector of the pooling operation, \(k_w\) is the size of the pooling window in the width direction, \(x_{j}\) is the element in \(j\)-th column of the input feature map, where\(j\) ranges from \(0\) to \(k_w - 1\). The Average pooling operates on the matrix generated after the LLM processes the tokens, i.e.,
\begin{align}
    y = \frac{1}{k_w}\sum_{j = 0}^{k_w - 1}x_{j},
\end{align}
\begin{figure}[!t]
\centering
\includegraphics[width=0.7\linewidth]{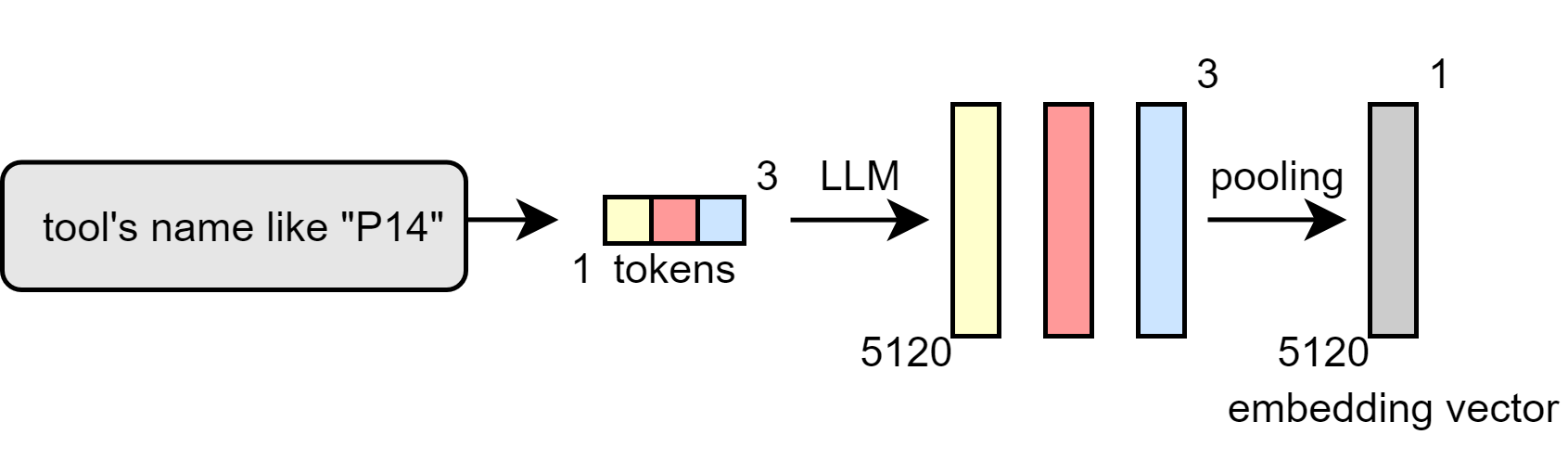}
\caption{An illustration of pooling operations on multiple token embeddings. 5120 is the word embedding dimension of LLaMA2-13B.}
\label{fig:Overviewofpooling}
\end{figure}
Average pooling selects the average value on the dimension of length(tokens) and generates a vector to provide information reference for the corresponding embedding vector. As the example shown in Figure~\ref{fig:Overviewofpooling}, when a tool is input, 3 tokens are correspondingly generated, and the dimension of the matrix generated by the LLM is [3, dim]. After calculating the average, it is transformed into a vector of [1, dim], dim refers to the dimension of model's hidden states of transformer. We apply the Max pooling to  selects the maximum value on the corresponding dimension, i.e.,
\begin{align}
    y=\max_{j\in [k_w-1]}x_{j}.
\end{align}

\subsection{Regularized Token Learning}

We initialize the learnable tool token embeddings $\mathbf{W}_{\tau}$ using the prior embeddings $\mathbf{W}^0_{\tau}$ from the previous subsection.  Subsequently, we update the learnable tool token embeddings $\mathbf{W}_{\tau}$ based on the next token prediction loss, while also ensure consistency with the LLM’s word embedding space by using the prior embeddings $\mathbf{W}^0_{\tau}$ as a reference. Therefore, our approach utilizes an overall loss function comprising two main components: the next-token prediction loss and a regularization term. These components are defined as follows:
\begin{align}
    \mathcal{L}(W_{\tau})&=\sum_{(s,s')\in D}\sum_{i = 1}^{N}-\log P(t_{i}'\vert t_{<i})\mathbb{I}_{t_{i}'\neq[\text{N/A}]} 
    +\lambda \left\|\mathbf{W}_{\tau}-\mathbf{W}_{\tau}^0\right\|_{2}^{2},
\end{align}
where the hyper-parameter $\lambda$ controls the trade-off between the next token prediction loss and  the regularization term that constrains the difference between $\textbf{W}_{\tau}$ and $\textbf{W}_{\tau}^0$. By minimizing the difference between $\mathbf{W}_{\tau}$ and $\mathbf{W}_{\tau}^0$ during optimization, we impose constraints on the learning process of $\mathbf{W}_{\tau}$.  

\begin{equation}
\mathcal{L}_{\text{reg}} = \lambda \left\|\mathbf{W}_{\tau}-\mathbf{W}_{\tau}^0\right\|_{2}^{2}.
\end{equation}

The $L_2$ regularization term $L_\text{reg}$ serves a crucial function during model training by imposing a constraint on the deviation between the learned tool token embeddings $\mathbf{W}_{\tau}$ and their corresponding prior embeddings $\textbf{W}_{\tau}^0$. This regularization mechanism ensures training stability by preventing excessive divergence of the model parameters from their initialization values. In the absence of such regularization, the model would be prone to overfitting the training data, consequently exhibiting suboptimal generalization performance on unseen data. The incorporation of $L_\text{reg}$ promotes more conservative parameter updates, thereby mitigating the risk of over-adaptation to noisy or idiosyncratic patterns present in the training set.

The regularization strength is governed by the hyperparameter $\lambda$, which determines the trade-off between model flexibility and constraint severity. Specifically,
For large values of $\lambda$, the optimization process strongly penalizes deviations from $\textbf{W}_{\tau}^0$, effectively anchoring the learned embeddings near their initial values. While this approach can reduce overfitting, it may excessively constrain the model's capacity to learn meaningful representations, potentially leading to underfitting and subpar performance on both training and test data.
Conversely, small values of $\lambda$ impose weak regularization constraints, permitting greater flexibility in parameter updates. However, this increased flexibility comes at the risk of overfitting, where the model may over-optimize to training set specifics at the expense of generalization capability.

This regularization framework establishes a critical balance between preserving prior knowledge (encoded in $\textbf{W}_{\tau}^0$ ) and adapting to new information during the learning process.

\section{Experiments}
In this section, we first introduce the datasets related to the three tasks: mathematical problem solving, knowledge-based question answering, and embodied plan generation. Then, we present the experimental results and conduct ablation studies.

\subsection{Datasets}
We consider four datasets including GSM8K-XL \cite{hao2024toolkengptaugmentingfrozenlanguage}, FuncQA \cite{hao2024toolkengptaugmentingfrozenlanguage}, KAMEL 
\cite{hao2024toolkengptaugmentingfrozenlanguage}, and VirtualHome \cite{hao2024toolkengptaugmentingfrozenlanguage}, specifically,

\textbf{GSM8K-XL:}
GSM8K-XL dataset\cite{hao2024toolkengptaugmentingfrozenlanguage} represents an enhanced version of the GSM8K \cite{cobbe2021trainingverifierssolvemath} benchmark, which consists of linguistically diverse grade school math word problems requiring sequential application of four basic arithmetic operations (+, −, ×, ÷) to reach final solutions. The original GSM8K dataset \cite{cobbe2021trainingverifierssolvemath} primarily contains problems with small numerical values, potentially limiting its effectiveness in evaluating contemporary large language models (LLMs) as it fails to sufficiently challenge their reasoning capacities or thoroughly examine their tool-utilization capabilities in complex problem-solving contexts \cite{bubeck2023sparksartificialgeneralintelligence}\cite{yao2023reactsynergizingreasoningacting}. To address this limitation, numerical values in the test set were systematically increased, resulting in the GSM8K-XL dataset comprising 568 test cases with substantially larger numbers to elevate computational difficulty.
The training process utilizes the original GSM8K training set with calculation annotations. From the available 6,054 examples, 5,054 serve as training data, while the remaining 1,000 function as validation samples. Thereby elevating computational complexity and enabling a more robust evaluation of LLMs' tool-assisted reasoning performance.

\textbf{FuncQA:} 
The FuncQA dataset \cite{hao2024toolkengptaugmentingfrozenlanguage} was developed to enhance the complexity of numerical reasoning tasks by evaluating models' ability to acquire and invoke appropriate tools when solving sophisticated mathematical problems involving multiple arithmetic operations and requiring multi-step reasoning \cite{hao2024toolkengptaugmentingfrozenlanguage}. This benchmark is designed to emulate realistic, computationally intensive scenarios that necessitate proficient utilization of diverse arithmetic tools, thereby imposing more stringent demands on models' tool-manipulation and logical reasoning capabilities.
The finalized FuncQA dataset comprises two distinct subsets: 68 one-hop questions that can be resolved through a single arithmetic operation, and 60 multi-hop questions requiring sequential reasoning steps. During the dataset construction process, a stratified sampling approach was employed - for each arithmetic operator, 47 training and 3 validation data points were systematically selected, culminating in a total of 611 training samples and 39 validation samples across all operators.

\textbf{KAMEL:}
Large language models (LLMs) often exhibit limitations in factual accuracy \cite{glaese2022improvingalignmentdialogueagents}, frequently generating erroneous or hallucinated content due to inherent knowledge constraints \cite{zha2023textalignmentefficientunified, Ji_2023, zha2023alignscoreevaluatingfactualconsistency,hao2023bertnetharvestingknowledgegraphs,yao2024retroformerretrospectivelargelanguage}. To mitigate these issues, knowledge base (KB) integration has emerged as a viable solution for reducing hallucination rates \cite{shuster2021retrievalaugmentationreduceshallucination,zong2024triadframeworkleveragingmultirole,huang-etal-2016-well}. In practical implementations, KB access is typically facilitated through application programming interfaces (APIs) \cite{talmor2018webknowledgebaseansweringcomplex}\cite{fu2020surveycomplexquestionanswering}, where each relational query can be conceptually framed as a tool operation \cite{qin2024toollearningfoundationmodels} - with subject entities as inputs and corresponding tail entities as outputs\cite{khattab2023demonstratesearchpredictcomposingretrievallanguage}.
The KAMEL\cite{hao2024toolkengptaugmentingfrozenlanguage} framework incorporates knowledge spanning 234 distinct Wikidata relations, with each relation mapped to a specific question template. For instance, the ``winner of" relation is associated with the template ``Who is the winner of [S]?", effectively converting Wikidata facts into queryable formats. This structure yields a total of 234 tool-like query mechanisms. To systematically investigate the relationship between tool quantity and model performance, we constructed four evaluation subsets through stratified sampling from the original test set. These subsets contain questions corresponding to 30, 60, 100, and 234 tools respectively, with each subset comprising 500 carefully curated questions.

\textbf{VirtualHome:}
Recent research has demonstrated significant interest in employing large language models (LLMs) as controllers for embodied agents \cite{huang2022languagemodelszeroshotplanners,singh2022progpromptgeneratingsituatedrobot,ahn2022icanisay,huang2023groundeddecodingguidingtext,xiang2023languagemodelsmeetworld}. While prompt-based approaches have achieved preliminary success \cite{yao2020calmexplorelanguagemodels}, significant challenges remain in enabling LLMs to develop comprehensive environmental understanding and generate grounded predictions. As highlighted by Mialon et al. \cite{mialon2023augmentedlanguagemodelssurvey}, LLMs demonstrate the capacity to utilize diverse tool types - including both information-gathering tools (e.g., mathematical or knowledge base tools) and physical-world interaction tools (e.g., embodied agent actions) - through fundamentally similar mechanisms.
VirtualHome \cite{puig2018virtualhomesimulatinghouseholdactivities} represents a foundational simulation platform and knowledge base for embodied intelligence research. This system, centered on common household activities, incorporates an ActivityPrograms knowledge base \cite{puig2018virtualhomesimulatinghouseholdactivities} containing numerous executable task plans. The dataset construction process involved selecting verbs and objects appearing with a minimum frequency threshold of 10 occurrences, resulting in a final configuration of 247 training tasks and 50 test tasks, encompassing 25 distinct verbs and 32 unique objects. When combined with the [END] function for plan termination, these elements collectively form 58 distinct tool tokens (toolkens) \cite{puig2018virtualhomesimulatinghouseholdactivities}.

\subsection{Implementation Details}

This section presents the application of our methodology across three well-defined domains exhibiting significant tool-utilization paradigms: 1) arithmetic operations for numerical reasoning tasks, 2) database API interactions for knowledge-based question answering, and 3) robotic action sequences for embodied planning generation. Our primary research objective focuses on enhancing large language models' (LLMs) capabilities in both precise tool prediction and effective tool application within the ToolkenGPT framework \cite{hao2024toolkengptaugmentingfrozenlanguage}. Regarding computational requirements, the training process was conducted using the following hardware configurations: the LLaMA2-7B model\cite{touvron2023llama2openfoundation} was trained on a single GeForce RTX 4090 GPU, the LLaMA2-13B\cite{touvron2023llama2openfoundation} implementation utilized two GeForce RTX 4090 GPUs, and the LLaMA2-70B model\cite{touvron2023llama2openfoundation} training was performed across eight GeForce RTX 4090D GPUs.

\subsubsection{\textbf{GSM8K-XL and FuncQA datasets}}

Building upon the methodology outlined in Section 3, we extract learning tokens corresponding to mathematical operation symbols to reinforce the constrained training of tool embeddings. During model inference, we employ a 4-shot Chain-of-Thought prompting strategy to augment the LLM's reasoning capabilities. Our comparative analysis incorporates the following baseline approaches. For the GSM8K-XL dataset, The toolken embeddings of learning tokens are trained with a subset of 5,063 examples. An additional 1,000 examples are reserved for validation. The embeddings were trained with a learning rate of 1e-3, performing early stopping based on the development set, with a maximum of 5 epochs. For the FuncQA dataset, the learning rate we use is 1e-4, and we perform early stopping based on the development set, with the maximal training epochs to be 10 epochs. The prompt of FuncQA is similar to the prompt of GSM8K-XL. we establish three principal baseline approaches:
\begin{itemize}
    \item \textbf{Chain-of-Thought (CoT)} \cite{wei2023chainofthoughtpromptingelicitsreasoning}: This state-of-the-art prompting technique employs carefully designed prompts to facilitate sequential reasoning during inference. We maintain consistency in reasoning chain examples across all comparative methods, including ToolkenGPT and TokenLearning implementations.
    \item \textbf{ReAct}~\cite{yao2023reactsynergizingreasoningacting}: An interactive paradigm that jointly generates reasoning traces and tool invocations. Our implementation adopts the specialized syntax for operator calls (e.g., $50 * 3.2 = <multiply>(50, 3.2) = 160$), where the system automatically triggers tool execution upon detecting this pattern during inference.
    \item \textbf{ToolkenGPT}~\cite{hao2024toolkengptaugmentingfrozenlanguage}: Our proposed approach represents tools as discrete tokens (``toolkens") embedded within the model's parameter space. When the generation process produces a toolken, the system automatically initiates the corresponding tool invocation.
\end{itemize}

For fair comparison, all methods utilize identical reasoning chain exemplars, varying only in their tool invocation syntax. We evaluate our approach using the LLaMA2 architecture at three different scales: LLaMA2-7B, LLaMA2-13B, and LLaMA2-70B models~\cite{touvron2023llama2openfoundation}.

\subsubsection{\textbf{KAMEL dataset}}
Toolken embeddings of learning tokens are trained with a learning rate of 1e-3, performing early stopping based on the development set, and trained for a maximum of 5 epochs. To rigorously assess our proposed methodology, we establish two principal baseline approaches on the KAMEL benchmark:
\begin{itemize}
\item \textbf{In-context Learning (ICL)}~\cite{qin2024toollearningfoundationmodels}: This paradigm represents a state-of-the-art approach for equipping LLMs with tool-usage capabilities through demonstration-based learning. Our implementation adopts a two-stage prompting strategy: (1) we first prepend the complete inventory of available tools along with their functional descriptions to the model's context window; (2) subsequently, we present the target query for processing. To mitigate the inherent constraints of limited context length in transformer-based architectures, we employ a space-optimized representation scheme where each tool is described using minimal lexical units (preferably single-word descriptors) without compromising operational semantics.
\item \textbf{ToolkenGPT}~\cite{hao2024toolkengptaugmentingfrozenlanguage}: Our proposed tokenized tool representation framework enables efficient tool composition through learned embeddings. The KAMEL dataset instantiation incorporates a comprehensive set of 234 distinct toolkens, each corresponding to a unique relational operation derived from the underlying knowledge graph. This representation allows for: (i) seamless integration with the model's existing vocabulary, (ii) efficient tool retrieval during inference, and (iii) scalable addition of new capabilities through token expansion.
\end{itemize}

Implementation Note: We employ constrained prompting techniques to restrict LLM outputs exclusively to relevant API calls, enabling precise evaluation of tool selection accuracy under controlled conditions.

\subsubsection{\textbf{VirtualHome dataset}}
Toolken embeddings of learning tokens are trained with a learning rate of 1e-4, performing early stopping based on the development set, with a maximum of 10 epochs. Note that all methods use the same prompts in this experiment. We establish parallel baseline methodologies for the VirtualHome environment to maintain consistent evaluation protocols:
\begin{itemize}
\item \textbf{In-context Learning (ICL)} \cite{qin2024toollearningfoundationmodels}: This approach implements a comprehensive priming strategy consisting of: (i) a complete enumeration of executable atomic actions, (ii) three exemplar task plans demonstrating proper tool sequencing, and (iii) the target task specification including its objective, operational parameters, and environmental context. This multi-component prompting architecture provides necessary grounding for situated action planning.
\item \textbf{ToolkenGPT} \cite{hao2024toolkengptaugmentingfrozenlanguage}: Our tokenized tool representation framework achieves efficient action composition through 58 discrete toolkens corresponding to: (a) 57 fundamental household actions, and (b) 1 termination token ([END]) for plan completion. Each toolken encapsulates both the semantic meaning and executable properties of its associated action.
\end{itemize}

In terms of computational resources, we train and test TokenLearning based on LLaMA2-7B, LLaMA2-13B and LLaMA2-70B using 1, 2 and 8 Nvidia RTX 4090 GPUs.

\subsection{Experimental Results}

\subsubsection{\textbf{GSM8K-XL}}
\begin{table*}[t]
  \centering
  \caption{Results on the GSM8K-XL with diferent models. For GSM8K-XL dataset, accuracy is evaluated based on an exact match (float numbers rounded to four decimals).
 }
    \begin{tabular}{cccc}
    \toprule
    \textbf{Methods} & LLaMA2-7B & LLaMA2-13B & LLaMA2-70B \\
    \midrule
    CoT \cite{wei2023chainofthoughtpromptingelicitsreasoning}   & 0.07  & 0.1267 & 0.3908 \\
    ReAct \cite{yao2023reactsynergizingreasoningacting} & 0.1461 & 0.2517 & 0.5123 \\
    ToolkenGPT \cite{hao2024toolkengptaugmentingfrozenlanguage} & 0.1397 & 0.2165 & 0.4789 \\
    TokenLearning (ours) & \textbf{0.1549} & \textbf{0.2852} & \textbf{0.5422} \\
    \bottomrule
    \end{tabular}%
  \label{Table:Exp:GSM8KXL}%
\end{table*}%
Table \ref{Table:Exp:GSM8KXL} presents a comprehensive evaluation of various methods on the GSM8K-XL dataset, revealing critical insights into large language models' mathematical reasoning capabilities. The Chain-of-Thought (CoT)\cite{wei2023chainofthoughtpromptingelicitsreasoning} approach demonstrates significant limitations, particularly in handling the dataset's extended numerical ranges, as it requires both precise mathematical-logical reasoning and accurate numerical computation - a well-documented challenge for pure LLM-based methods. This computational bottleneck becomes increasingly pronounced with larger numerical values in the GSM8K-XL benchmark. In contrast, tool-augmented methods including ReAct\cite{yao2023reactsynergizingreasoningacting}, ToolkenGPT\cite{hao2024toolkengptaugmentingfrozenlanguage}, and our proposed TokenLearning approach achieve substantially improved performance by externalizing numerical operations, thereby ensuring correct computational results when the model's reasoning process is valid. Notably, our TokenLearning method, building upon ToolkenGPT's framework, delivers consistent performance gains of approximately 3\% across model sizes. While ReAct demonstrates strong results on the LLaMA2-70B model (51.23\%), highlighting the enhanced comprehension capabilities of larger-scale models, our TokenLearning approach ultimately achieves superior performance (54.22\%), demonstrating that specialized training methodologies can further optimize model capabilities even when applied to already proficient large-scale architectures.

\begin{table*}[htbp]
  \centering
  \caption{Results on FuncQA dataset in different methods on LLaMA2-70B model under multi-hops and one-hop. 
}
    \begin{tabular}{lcccc}
    \toprule
    \textbf{Method} & \textbf{CoT} \cite{wei2023chainofthoughtpromptingelicitsreasoning} & \textbf{ReAct} \cite{yao2023reactsynergizingreasoningacting} & \textbf{ToolkenGPT }\cite{hao2024toolkengptaugmentingfrozenlanguage} & \textbf{TokenLearning (ours)} \\
    \midrule
    Multi-Hops & 0.06 & 0.176 & 0.147 & \textbf{0.162} \\
    One-Hop & 0.25 & 0.38 & 0.6 & \textbf{0.65} \\
    \bottomrule
    \end{tabular}%
  \label{Table:Exp:FuncQA}%
\end{table*}%

Note that for  FuncQA (One-Hop) dataset, accuracy is evaluated based on an exact match (float numbers rounded to three decimals). In FuncQA (Multi-Hops), we allow a margin of error of 0.1\% to account for potential errors at each step of Multi-Hops reasoning.
As presented in Table \ref{Table:Exp:FuncQA}, our TokenLearning method achieves superior performance on the One-Hop task with 0.65 accuracy, significantly outperforming all baseline approaches on the LLaMa2-70B model. For Multi-Hop reasoning, while our method demonstrates a marked improvement (0.162) over ToolkenGPT (0.147), it remains marginally inferior to ReAct (0.176). These results suggest that while learned tool representations exhibit strong performance in simpler one-hop scenarios, their effectiveness in complex multi-hop reasoning may be constrained by the precision of token-level representations when training data is limited. Notably, ReAct's superior multi-hop performance underscores the remarkable capability of large language models to dynamically select appropriate tools through well-designed prompting and in-context learning, even without explicit tool token training, highlighting the complementary advantages of prompt-based versus learned tool invocation mechanisms in different reasoning contexts.

\subsubsection{\textbf{Kamel}}

\begin{figure}[t]
\centering
\includegraphics[width=0.6\linewidth]{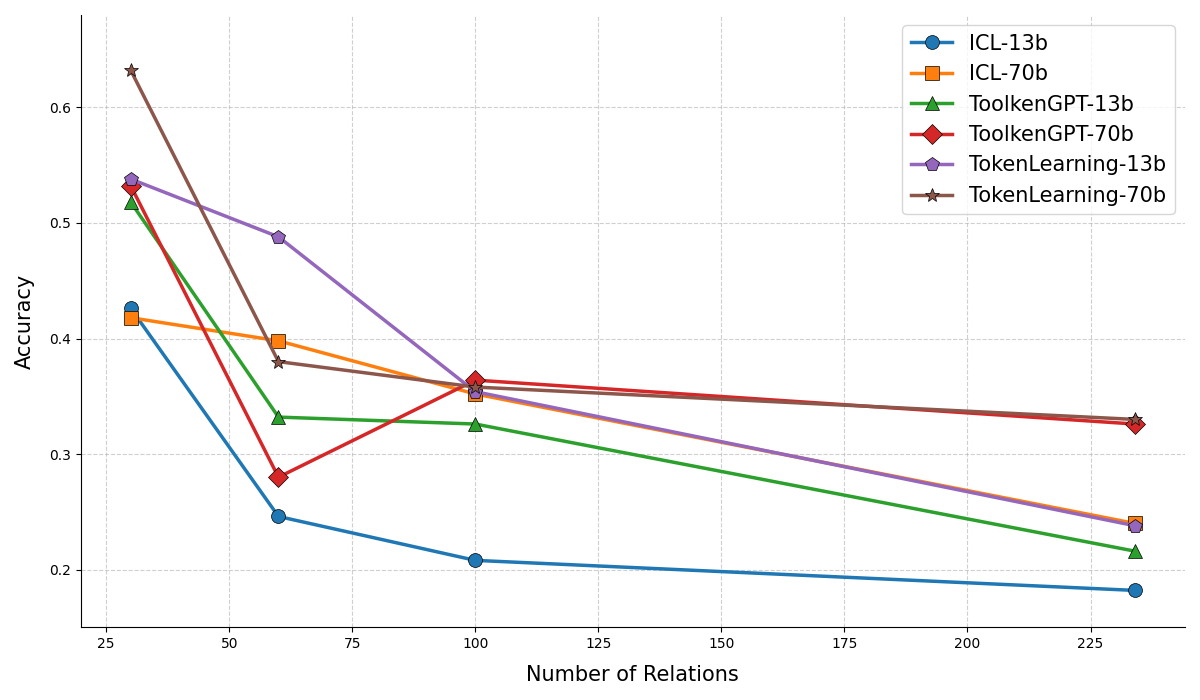}
\caption{Performance of Tokenlearning and baselines on 4 testsets(each testset consists of questions related to different numbers of relations, corresponding to 30, 60, 100, and 234, respectively, the size of each testset is 500) involving different numbers of tools (relations) from KAMEL. 
}
\label{fig:ResultsOnKAMEL}
\end{figure}

Our experimental evaluation across four test sets with varying relations demonstrates distinct performance characteristics among the compared approaches, as illustrated in Figure \ref{fig:ResultsOnKAMEL}. The in-context learning (ICL) methods exhibit notable limitations in tool selection accuracy, with both ICL-13b and ICL-70b variants showing significantly lower performance compared to tool-augmented approaches. Notably, our TokenLearning method achieves consistent improvements of approximately 3\% or greater over ToolkenGPT across all test sets and model scales (LLaMA2-13B and LLaMA2-70B), with the most substantial gains observed in the LLaMA2-70B configurations. These results substantiate that our learned token representations maintain effective guidance for tool selection despite the inherent challenge of API relations being composed of semantically irrelevant tokens, highlighting the robustness of our approach in capturing functional relationships beyond surface-level token semantics. The progressive performance enhancement from ICL to ToolkenGPT and further to TokenLearning suggests a clear hierarchy in tool utilization effectiveness, with our method establishing a new state-of-the-art in tool-augmented language model performance.

\subsubsection{\textbf{VirtualHome}}
The experimental results presented in Table \ref{Table:Exp:VirtualHome} demonstrate that our TokenLearning method achieves consistent performance improvements, delivering approximately 2\% higher accuracy than ToolkenGPT (0.72 vs 0.68 for LLaMa2-13B and 0.78 vs 0.76 for LLaMa2-70B models) while significantly outperforming In-Context Learning (ICL) by substantial margins (0.72 vs 0.24 for LLaMa2-13B and 0.78 vs 0.34 for LLaMa2-70B), thereby establishing a new state-of-the-art for tool-augmented task performance on the VirtualHome benchmark across both model scales.

\begin{table}[t]
  \centering
  \caption{Performance comparison on VirtualHome dataset across LLaMA2 model sizes\cite{touvron2023llama2openfoundation}. Success accuracy measures the proportion of scripts achieving correct final states.}
    \begin{tabular}{lccc}
    \toprule
    \textbf{Approach} & \textbf{ICL\cite{qin2024toollearningfoundationmodels}} & \textbf{ToolkenGPT\cite{hao2024toolkengptaugmentingfrozenlanguage}} & \textbf{ Ours} \\
    \midrule
    LLaMa2-13B & 0.24 & 0.68 & \textbf{0.72} \\
    LLaMa2-70B & 0.34 & 0.76 & \textbf{0.78} \\
    \bottomrule
    \end{tabular}%
  \label{Table:Exp:VirtualHome}%
\end{table}%

\subsection{Ablation Studies }
\subsubsection{Initialization}
\begin{table}[t]
  \centering
  \caption{Results on of ablation study GSM8K - XL dataset in different initialization on LLaMA2-70b. Accuracy is evaluated based on an exact match (float numbers rounded to four decimals).}
    \begin{tabular}{ccc}
    \toprule
    \textbf{Initialization}  & \textbf{Irrelevant vocabulary} & \textbf{Tools' name }\\
    \midrule
    LLaMA2-70B & 0.4526  & 0.4683 \\
    LLaMA2-13B & 0.2306  & 0.2324 \\
    \bottomrule
    \end{tabular}%
  \label{init:gsm8k}%
\end{table}%

\begin{table}[t]
  \centering
  \caption{Results of ablation study on VirtualHome dataset in different initialization on LLaMA2-13b and LLaMA2-70b. Success accuracy is a relaxed variant meaning the proportion of scripts that have reached the correct final state, but not necessarily ending with it.
}
    \begin{tabular}{lcc}
    \toprule
    \textbf{Initialization} & LLaMA2-13B & LLaMA2-70B \\
    \midrule
    Maximum pooling & 0.66 & 0.74 \\
    Average pooling & 0.72 & 0.76 \\
    \bottomrule
    \end{tabular}%
  \label{init:virtualhome}%
\end{table}%
The quality of the additional matrix \(W_{\tau}\) is significantly influenced by different initialization methods. In the specific context of the GSM8K-XL dataset, tokens corresponding to fundamental arithmetic operations (addition, subtraction, multiplication, and division) exhibit distinct and well-defined directional properties. Building upon this observation, we carefully designed and conducted a series of experiments to comprehensively evaluate the model's performance across various input tokens, including semantically neutral tokens such as ``one'', ``two'', ``three'', and ``four''. This experimental design enables clear demonstration of the directional relationship between learning tokens and tool tokens through comparative analysis.
As evident from the results presented in Table \ref{init:gsm8k}, model accuracy is notably compromised when irrelevant tokens are employed for initialization, compared to the performance achieved using our proposed method. These empirical results robustly validate the superiority of our initialization approach. Notably, our method maintains high accuracy even under conditions of relatively limited training data. Furthermore, these findings corroborate that \(\textbf{W}_{\tau}^0\) exhibits stronger directional guidance, enabling more effective steering of the model along the desired learning trajectory.

To further investigate the impact of pooling operations on model initialization, we conducted additional experiments. Tables \ref{init:virtualhome}  present comparative analyses of initialization performance using matrices generated through different pooling operations. Our detailed examination reveals that average pooling, when applied after comprehensive integration of all token information, demonstrates particularly pronounced directional properties. This finding offers valuable strategic insights for model initialization approaches.

\begin{table*}[t]
  \centering
  \caption{Results of ablation study on GSM8K-XL, FuncQA-oh, KAMEL and VirtualHome datasets in different regularization term constraint coefficient \(\lambda\) on LLaMA2-70B and LLaMA2-13B.
  }
\begin{tabular}{ccrrrrrrc}
    \toprule
    \textbf{Dataset} & \textbf{Model} & \multicolumn{1}{c}{$\lambda$=1e-6} & \multicolumn{1}{c}{$\lambda$=1e-5} & \multicolumn{1}{c}{$\lambda$=1e-4} & \multicolumn{1}{c}{$\lambda$=1e-3} & \multicolumn{1}{c}{$\lambda$=1e-2} & \multicolumn{1}{c}{$\lambda$=2e-2} & \textbf{Pooling} \\
    \midrule
    \multirow{2}[2]{*}{GSM8K-XL\cite{hao2024toolkengptaugmentingfrozenlanguage}} & LLaMA2-13B & \multicolumn{1}{c}{0.2253} & \multicolumn{1}{c}{0.2324} & \multicolumn{1}{c}{0.2430} & \multicolumn{1}{c}{\textbf{0.2869}} & \multicolumn{1}{c}{0.2852} & \multicolumn{1}{c}{0.2799} & - \\
          & LLaMA2-70B & \multicolumn{1}{c}{0.4982} & \multicolumn{1}{c}{0.4929} & \multicolumn{1}{c}{0.4701} & \multicolumn{1}{c}{0.5404} & \multicolumn{1}{c}{\textbf{0.5422}} & \multicolumn{1}{c}{0.5352} & - \\
    \midrule
    \multirow{4}[4]{*}{FuncQA-oh\cite{hao2024toolkengptaugmentingfrozenlanguage}} & \multirow{2}[2]{*}{LLaMA2-13B} & \multicolumn{1}{c}{0.55} & \multicolumn{1}{c}{0.55} & \multicolumn{1}{c}{0.53} & \multicolumn{1}{c}{0.53} & \multicolumn{1}{c}{0.48} & \multicolumn{1}{c}{0.45} & Max \\
          &       & \multicolumn{1}{c}{0.55} & \multicolumn{1}{c}{0.55} & \multicolumn{1}{c}{0.53} & \multicolumn{1}{c}{0.53} & \multicolumn{1}{c}{0.45} & \multicolumn{1}{c}{0.45} & Avg \\
\cmidrule{2-2}          & \multirow{2}[2]{*}{LLaMA2-70B} & \multicolumn{1}{c}{0.63} & \multicolumn{1}{c}{0.63} &   \multicolumn{1}{c}{0.63}    & \multicolumn{1}{c}{0.65} & \multicolumn{1}{c}{0.58} & \multicolumn{1}{c}{0.55} & Max \\
          &       & \multicolumn{1}{c}{0.63} & \multicolumn{1}{c}{0.63} &   \multicolumn{1}{c}{0.63}    & \multicolumn{1}{c}{\textbf{0.65}}      & \multicolumn{1}{c}{0.58} & \multicolumn{1}{c}{0.55} & Avg \\
    \midrule
    \multirow{10}[8]{*}{KAMEL\cite{hao2024toolkengptaugmentingfrozenlanguage}} & \multirow{2}[4]{*}{Tool'num} & \multicolumn{1}{c}{$\lambda$=1e-6} & \multicolumn{1}{c}{$\lambda$=1e-5} & \multicolumn{1}{c}{$\lambda$=1e-4} & \multicolumn{1}{c}{$\lambda$=1e-4} & \multicolumn{1}{c}{$\lambda$=1e-3} & \multicolumn{1}{c}{$\lambda$=1e-2} & \multirow{2}[4]{*}{} \\
\cmidrule{3-8}          &       & \multicolumn{3}{c}{LLaMA2-13B} & \multicolumn{3}{c}{LLaMA2-70B} &  \\
\cmidrule{2-8}          & 30    & \multicolumn{1}{c}{0.512} & \multicolumn{1}{c}{0.538} & \multicolumn{1}{c}{0.632} & \multicolumn{1}{c}{0.592} & \multicolumn{1}{c}{0.632} & \multicolumn{1}{c}{0.592} & \multirow{4}[2]{*}{Avg} \\
          & 60    & \multicolumn{1}{c}{0.466} & \multicolumn{1}{c}{0.488} & \multicolumn{1}{c}{0.226} & \multicolumn{1}{c}{0.414} & \multicolumn{1}{c}{0.38} & \multicolumn{1}{c}{0.348} &  \\
          & 100   & \multicolumn{1}{c}{0.388} & \multicolumn{1}{c}{0.354} & \multicolumn{1}{c}{0.198} & \multicolumn{1}{c}{0.292} & \multicolumn{1}{c}{0.358} & \multicolumn{1}{c}{0.282} &  \\
          & 234   & \multicolumn{1}{c}{0.254} & \multicolumn{1}{c}{0.238} & \multicolumn{1}{c}{0.17} & \multicolumn{1}{c}{0.236} & \multicolumn{1}{c}{0.33} & \multicolumn{1}{c}{0.238} &  \\
\cmidrule{2-9}          & 30    & \multicolumn{1}{c}{0.38} & \multicolumn{1}{c}{0.498} & \multicolumn{1}{c}{0.592} & \multicolumn{1}{c}{0.564} & \multicolumn{1}{c}{0.592} & \multicolumn{1}{c}{0.588} & \multirow{4}[2]{*}{Max} \\
          & 60    & \multicolumn{1}{c}{0.268} & \multicolumn{1}{c}{0.348} & \multicolumn{1}{c}{0.132} & \multicolumn{1}{c}{0.256} & \multicolumn{1}{c}{0.246} & \multicolumn{1}{c}{0.262} &  \\
          & 100   & \multicolumn{1}{c}{0.222} & \multicolumn{1}{c}{0.18} & \multicolumn{1}{c}{0.112} & \multicolumn{1}{c}{0.36} & \multicolumn{1}{c}{0.318} & \multicolumn{1}{c}{0.38} &  \\
          & 234   & \multicolumn{1}{c}{0.192} & \multicolumn{1}{c}{0.154} & \multicolumn{1}{c}{0.102} & \multicolumn{1}{c}{0.322} & \multicolumn{1}{c}{0.314} & \multicolumn{1}{c}{0.304} &  \\
    \midrule
    \multirow{5}[6]{*}{VirtualHome\cite{hao2024toolkengptaugmentingfrozenlanguage}} & Model & \multicolumn{1}{c}{$\lambda$=1e-3} & \multicolumn{1}{c}{$\lambda$=1e-2} & \multicolumn{1}{c}{$\lambda$=0.1} & \multicolumn{1}{c}{$\lambda$=0.8} & \multicolumn{1}{c}{$\lambda$=0.9} & \multicolumn{1}{c}{$\lambda$=1.0} &  \\
\cmidrule{2-8}          & \multirow{2}[2]{*}{LLaMA2-13B} & \multicolumn{1}{c}{0.70} & \multicolumn{1}{c}{0.72} & \multicolumn{1}{c}{0.68} & \multicolumn{1}{c}{0.72} & \multicolumn{1}{c}{\textbf{0.72}} & \multicolumn{1}{c}{0.70} & Avg \\
          &   &   \multicolumn{1}{c}{0.68}    &   \multicolumn{1}{c}{0.66}  &  \multicolumn{1}{c}{0.66}  &  \multicolumn{1}{c}{0.70}   &   \multicolumn{1}{c}{0.66} & \multicolumn{1}{c}{0.68}  & Max \\
\cmidrule{2-2}          & \multirow{2}[2]{*}{LLaMA2-70B} & \multicolumn{1}{c}{\textbf{0.78}} & \multicolumn{1}{c}{0.76} & \multicolumn{1}{c}{0.76} & \multicolumn{1}{c}{0.74} & \multicolumn{1}{c}{0.76} & \multicolumn{1}{c}{0.68} & Avg \\
          &       & \multicolumn{1}{c}{0.72} & \multicolumn{1}{c}{0.74} & \multicolumn{1}{c}{0.78} & \multicolumn{1}{c}{0.78} & \multicolumn{1}{c}{0.72} & \multicolumn{1}{c}{0.74} & Max \\
    \bottomrule
    \end{tabular}%
  \label{tab:total}%
\end{table*}

\subsubsection{Pooling}

\begin{table*}
  \centering
  \caption{Results of ablation study on KAMEL dataset in different pooling operations on LLaMA2-13b. Accuracy is evaluated based on an exact match (float numbers rounded to three decimals).}
    \begin{tabular}{ccccc}
    \toprule
    \textbf{Tool’num} & Max($\lambda=0.001$) & Max($\lambda=0.01$) & Avg($\lambda=0.01$) & Avg($\lambda=0.001$) \\
    \midrule
    30 & 0.498 & 0.592 & \textbf{0.632} & 0.538 \\
    60 & \textbf{0.348} & 0.132 & 0.226 & 0.188 \\
    100 & 0.18 & 0.112 & 0.198 & \textbf{0.354} \\
    234 & 0.154 & 0.102 & 0.17 & \textbf{0.238} \\
    \bottomrule
    \end{tabular}
  \label{Table:Exp:pooling}%
\end{table*}%

In our experiments, we investigated the influence of different pooling operations on model performance. On the KAMEL dataset, owing to the uniqueness of its relations (where no separate relevant tokens exist in the vocabulary), the choice of pooling operation significantly affects the learning of tokens. We evaluated two distinct approaches—max pooling and average pooling—and observed a pronounced divergence in the accuracy of the generated outputs from the Large Language Model (LLM).

As illustrated in Table \ref{Table:Exp:pooling}, which compares max pooling and average pooling under varying constraint term coefficients on the LLaMA2-13B model with the KAMEL dataset, average pooling yields more stable and consistent results than max pooling. Specifically, as the number of tools (relations) increases, the accuracy of max pooling degrades sharply, whereas average pooling maintains robust performance. Furthermore, when encountering unseen ``toolkens'', superior results are achieved by holistically leveraging the LLM’s contextual information to derive the learning ``tokens''.
These findings suggest that average pooling offers greater stability and generalizability, particularly in scenarios involving an expanding set of tools or unfamiliar ``toolkens''.

\subsubsection{Regularization}

Regarding tool invocation accuracy, our analysis demonstrates that an appropriately tuned regularization coefficient \(\lambda\) consistently enhances performance. This parameter effectively guides the model in learning tool invocation patterns while leveraging prior knowledge to constrain updates of the tool-related parameter matrix \(\textbf{W}_{\tau}\), thereby enabling precise generation of tool tokens for successful invocations.
However, suboptimal selection of \(\lambda\) adversely impacts accuracy. An excessively large \(\lambda\) imposes overly stringent constraints, preventing the model from adequately learning the essential features required for tool invocation and consequently impairing its adaptability to novel tools or evolving usage contexts. Conversely, an insufficient \(\lambda\) value predisposes the model to overfitting, which manifests as degraded tool invocation accuracy in complex real-world scenarios.

We systematically investigate the effect of the constraint term coefficient \(\lambda\) on model performance through comprehensive experimentation. To thoroughly evaluate this relationship, we conduct extensive testing across three distinct datasets while varying the values of \(\lambda\). This empirical analysis provides quantitative insights into how different regularization strengths influence the system's behavior under controlled experimental conditions.

As shown in the tables \ref{tab:total}, the results of the LLaMA2-13B and LLaMA2-70B models under different constraint term coefficients $\lambda$ and the number of tools (for the KAMEL dataset) on the GSM8K-XL, KAMEL, and VirtualHome datasets are presented. The trends of the results of each model with the change of the constraint term coefficient $\lambda$ vary on different datasets.

Tables \ref{tab:total}  \footnote{For the GSM8K-XL dataset, Accuracy is evaluated based on an exact match (float numbers rounded to four decimals). 
  For the FuncQA-oh dataset, Accuracy is evaluated based on an exact match (float numbers rounded to two decimals). For the KAMEL dataset, Accuracy is evaluated based on an exact match (float numbers rounded to three decimals). 
  For the VirtualHome dataset, success of accuracy is a relaxed variant meaning the proportion of scripts that have reached the correct final state, but not necessarily ending with it, accuracy is evaluated based on an exact match (float numbers rounded to two decimals).} present the comprehensive evaluation results of LLaMA2-13B and LLaMA2-70B models across three benchmark datasets: GSM8K-XL, KAMEL, and VirtualHome. The experimental data systematically demonstrate model performance under varying constraint term coefficients \(\lambda\), with additional analysis of tool quantity variations specific to the KAMEL dataset.
Our comparative analysis reveals significant dataset-dependent variations in how each model's performance metrics evolve with changes in the regularization strength \(\lambda\). These differential response patterns highlight the context-sensitive nature of optimal \(\lambda\) selection across different problem domains and loss function.

\section{Conclusion}

In this work, we propose TokenLearning, a novel methodology designed to augment the tool-utilization capabilities of large language models (LLMs). TokenLearning employs LLMs to preprocess tokens associated with ``toolkens'', subsequently constructing an embedding matrix through pooling and concatenation operations. This matrix serves a dual purpose: 1) as an initialization mechanism, and 2) as a constraint during the training process, thereby improving the directional specificity of ``toolkens'' while enhancing both the accuracy and adaptability of tool invocations.

To evaluate the efficacy of TokenLearning, we conducted extensive experiments across a diverse set of tasks, including numerical reasoning, knowledge-based question answering, and embodied plan generation. Our results demonstrate that TokenLearning significantly enhances LLM performance by enabling more effective mastery of newly introduced ``toolkens''. Specifically, the proposed approach facilitates more precise tool prediction and utilization, exhibiting not only high accuracy in tool invocation but also robust adaptability to previously unseen tools.
These findings suggest that TokenLearning provides a promising framework for advancing LLM development in the domain of tool integration, opening new avenues for future research in this direction.

\section{Appendix}
\appendix
\section{Details of Experiments}

In this section, we describe the training configuration. Regarding the prompts used for different methods, they remain consistent with those in ToolkenGPT \cite{hao2024toolkengptaugmentingfrozenlanguage}.
\subsection{Details of numerical reasoning [GSM8K-XL]}

Prompt for\textbf{ Chain of Thought (CoT)} and ToolkenGPT reasoning mode:

\begin{tcbraster}[
    raster columns=1, 
    raster equal height=rows, 
    raster left skip=0pt, raster right skip=0pt, 
    colback=orange!20!white,
    colframe=orange,
    boxrule=0.5pt,
    sharp corners, 
    raster row skip=0pt 
]
    \begin{tcolorbox}[
        title={\textbf{Prompt of Chain of Thought (CoT)}},
        fonttitle=\bfseries,
        coltitle=black, 
        toprule=1.5pt, 
        bottomrule=0pt 
    ]
    \end{tcolorbox}

    \begin{tcolorbox}[
        colback=orange!10!white,
        toprule=0pt, bottomrule=0pt 
    ]
    Answer the following questions step by step.
    \end{tcolorbox}

    \begin{tcolorbox}[
        title={\textbf{Examples}},
        fonttitle=\bfseries,
        coltitle=black, 
        toprule=0pt, 
        bottomrule=0pt 
    ]
    \end{tcolorbox}

    \begin{tcolorbox}[
        breakable,
        colback=orange!10!white,
        toprule=0pt, 
        bottomrule=1.5pt 
    ]
    \begin{itemize}
        \item \textbf{Question 1}: Mark has 3 tanks for pregnant fish. Each tank has 4 pregnant fish and each fish gives birth to 20 young. How many young fish does he have at the end?
    
    \textbf{Answer 1}:  He has $4\times3 = 12$ pregnant fish.  They give birth to $12\times20=240$ fish. \#\#\# 240

    \item 
    \textbf{Question 2:} The math questions in a contest are divided into three rounds: easy, average, and hard. There are corresponding points given for each round. That is 2, 3, and 5 points for every correct answer in the easy, average, and hard rounds, respectively. 
    Suppose Kim got 6 correct answers in the easy; 2 correct answers in the average; and 4 correct answers in the difficult round, what are her total points in the contest?

    \textbf{Answer 2:}   Kim got $6\times2 = 12$ points in the easy round. 
      She got $2\times3 = 6$ points in the average round. 
       She got $4\times5 = 20$ points in the difficult round. 
        So her total points is $12 + 6+20 = 38$ points. \#\#\# 38
\item 
    \textbf{Question 3: }A clothing store sells 20 shirts and 10 pairs of jeans. A shirt costs $\$10$ each and a pair of jeans costs twice as much. How much will the clothing store earn if all shirts and jeans are sold?

    \textbf{Answer 3:}   
    Twenty shirts amount to $\$10\times20=\$200$. 
    The cost of each pair of jeans is $\$10\times2=\$20$. 
    So 10 pairs of jeans amount to $\$20\times10 = \$200$. 
    Therefore, the store will earn $\$200+\$200=\$400$ if all shirts and jeans are sold. \#\#\# 400
\item 
    \textbf{Question 4:} Arnold's collagen powder has 18 grams of protein for every 2 scoops. His protein powder has 21 grams of protein per scoop. And his steak has 56 grams of protein. If he has 1 scoop of collagen powder, 1 scoop of protein powder and his steak, how many grams of protein will he consume?
\item
    \textbf{Answer 4:} 2 scoops of collagen powder have 18 grams of protein and he only has 1 scoop so he consumes $18\div2 = 9$ grams of protein. He has 9 grams collagen powder, 21 grams of protein powder and 56 grams in his steak for a total of $9 + 21+56=86$ grams of protein. \#\#\# 86

\item \textbf{Question: [QUESTION]}

\textbf{Answer:}
\end{itemize}
\end{tcolorbox}
\end{tcbraster}
Prompt for ReAct, answer the following questions with \textbf{<add>},\textbf{ <subtract>}, \textbf{<multiply>}, \textbf{<divide>} operators:

\begin{tcbraster}[
    raster columns=1, 
    raster equal height=rows, 
    raster left skip=0pt, raster right skip=0pt, 
    colback=orange!20!white,
    colframe=orange,
    boxrule=0.5pt,
    sharp corners, 
    raster row skip=0pt 
]
    \begin{tcolorbox}[
        title={\textbf{Prompt of Chain of Thought (CoT) under operators}},
        fonttitle=\bfseries,
        coltitle=black, 
        toprule=1.5pt, 
        bottomrule=0pt 
    ]
    \end{tcolorbox}

    \begin{tcolorbox}[
        colback=orange!10!white,
        toprule=0pt, bottomrule=0pt 
    ]
    Answer the following questions with \textit{<add>}, \textit{<subtract>}, \textit{<multiply>}, \textit{<divide>} operators
    \end{tcolorbox}

    \begin{tcolorbox}[
        title={\textbf{Examples}},
        fonttitle=\bfseries,
        coltitle=black, 
        toprule=0pt, 
        bottomrule=0pt 
    ]
    \end{tcolorbox}

    \begin{tcolorbox}[
        breakable,
        colback=orange!10!white,
        toprule=0pt, 
        bottomrule=1.5pt 
    ]
\begin{itemize}
\item 
    \textbf{Question 1:} Mark has 3 tanks for pregnant fish.  Each tank has 4 pregnant fish and each fish gives birth to 20 young.  How many young fish does he have at the end?
    
    \textbf{Answer 1:} He has \(4\times3 = \textit{<multiply>}(4, 3)=12\) pregnant fish. They give birth to \(12\times20=\textit{<multiply>}(12, 20)=240\) fish. \#\#\# 240
\item     
    \textbf{Question 2:} The math questions in a contest are divided into three rounds: easy, average, and hard. There are corresponding points given for each round. That is 2, 3, and 5 points for every correct answer in the easy, average, and hard rounds, respectively. Suppose Kim got 6 correct answers in the easy; 2 correct answers in the average; and 4 correct answers in the difficult round, what are her total points in the contest?
    
    \textbf{Answer 2:} Kim got \(6\)\(\times2\) \( = \textit{<multiply>}(6, 2)=12\) points in the easy round. She got \(2\)\(\times3\) \( = \textit{<multiply>}(2, 3)=6\) points in the average round. She got \(4\)\(\times5\) \( = \textit{<multiply>}(4, 5)=20\) points in the difficult round. So her total points is \(12\) \(+6\)  \(+20\) \(= \textit{<add>}(12, 6, 20)=38\) points. \#\#\# 38
\item 
    \textbf{Question 3:} A clothing store sells 20 shirts and 10 pairs of jeans. A shirt costs \(\$10\) each and a pair of jeans costs twice as much. How much will the clothing store earn if all shirts and jeans are sold?

    \textbf{Answer 3:} Twenty shirts amount to \(\$10\times20=\$\textit{<multiply>}(10, 20)=200\). The cost of each pair of jeans is \(\$10\times2=\$\textit{<multiply>}(10, 2)=20\). So 10 pairs of jeans amount to \(\$20\times10=\$\textit{<multiply>}(20, 10)=200\). Therefore, the store will earn \(\$200+\$200=\$\textit{<add>}(200, 200)=400\) if all shirts and jeans are sold. \#\#\# 400
\item 
    \textbf{Question 4:} Arnold's collagen powder has 18 grams of protein for every 2 scoops.  His protein powder has 21 grams of protein per scoop.  And his steak has 56 grams of protein.   If he has 1 scoop of collagen powder, 1 scoop of protein powder and his steak, how many grams of protein will he consume?
    
    \textbf{Answer 4:} 2 scoops of collagen powder have 18 grams of protein and he only has 1 scoop so he consumes \(18\div2= \textit{<divide>} (18, 2)=9\) grams of protein.He has 9 grams collagen powder, 21 grams of protein powder and 56 grams in his steak for a total of \(9 + 21+56=\textit{<add>}(9, 21, 56)=86\) grams of protein. \#\#\# 86
    
\item  \textbf{Question: [QUESTION]}

\textbf{Answer:}
\end{itemize}
\end{tcolorbox}
\end{tcbraster}
\subsection{Details of Embodied Plan Generation[VirtualHome]}
Below are the prompts used for LLMs to generate plans. Note that all methods employ identical prompts in this experiment.

\begin{tcbraster}[
    raster columns=1, 
    raster equal height=rows, 
    raster left skip=0pt, raster right skip=0pt, 
    colback=orange!20!white,
    colframe=orange,
    boxrule=0.5pt,
    sharp corners, 
    raster row skip=0pt 
]
    \begin{tcolorbox}[
        title={\textbf{Background Description}},
        fonttitle=\bfseries,
        coltitle=black, 
        toprule=1.5pt, 
        bottomrule=0pt 
    ]
    \end{tcolorbox}

    \begin{tcolorbox}[
        colback=orange!10!white,
        toprule=0pt, bottomrule=0pt 
    ]
    I am a household robot and I can take actions from `[FIND]’, `[SIT]’,
`[SWITCHON]’, `[TURNTO]’, `[LOOKAT]’, `[TYPE]’, `[WALK]’, `[LIE]’, `[GRAB]’, `[READ]’, `[WATCH]’, `[POINTAT]’, `[TOUCH]’, `[SWITCHOFF]’, `[OPEN]’, `[PUSH]’, `[PUTOBJBACK]’, `[CLOSE]’, `[DRINK]’, `[RUN]’, `[DROP]’, `[PULL]’.
    \end{tcolorbox}

    \begin{tcolorbox}[
        title={\textbf{Examples}},
        fonttitle=\bfseries,
        coltitle=black, 
        toprule=0pt, 
        bottomrule=0pt 
    ]
    \end{tcolorbox}

    \begin{tcolorbox}[
        breakable,
        colback=orange!10!white,
        toprule=0pt, 
        bottomrule=1.5pt 
    ]
\begin{itemize}
\item 
\textbf{Task 1:}
I am in [`bathroom’]. The objects I can manipulate are [`faucet’, `keyboard’, `television’, `coffe\_maker’, `chair’, `button’, `pillow’, `phone’, `cup’, `couch’, `freezer’, `desk’, `oven’, `light’, `table’, `bedroom’, `dining\_room’, `cupboard’, `computer’, `sink’, `mail’, `bed’, `mouse’, `home\_office’].

\textbf{Goal:}
Write an email

\textbf{Hint:}
I went near the computer and turned it on, then sent the mail

\textbf{Plan:}

\quad [WALK] <home\_office>

\quad [WALK] <table>

\quad [FIND] <table>

\quad [WALK] <table>

\quad [FIND] <computer>

\quad [TURNTO] <computer>

\quad [LOOKAT] <computer>

\quad [TURNTO] <computer>

\quad [SWITCHON] <computer>

\quad [FIND] <mail>

\quad [TURNTO] <mail>

\item 
\textbf{Task 2:}
I am in [`home\_office’]. The objects I can manipulate are [`faucet’,
`novel’, `keyboard’, `television’, `newspaper’, `chair’, `coffe\_maker’,
`pillow’, `phone’, `check’, `couch’, `freezer’, `desk’, `toothbrush’,
`oven’, `light’, `food\_food’, `table’, `bookmark’, `bedroom’,
`dining\_room’, `computer’, `sink’, `mail’, `bed’, `cat’, `mouse’,
`home\_office’, `pot’].

\textbf{Goal:}
Work

\textbf{Hint:}
Find the computer. Turn it on by pressing the on button. Wait for it to
load. Use the mouse and keyboard to perform your tasks on screen.

\textbf{Plan:}

\quad [FIND] <computer>

\quad [SWITCHON] <computer>

\quad [FIND] <mouse>

\quad [TOUCH] <mouse>

\quad [FIND] <keyboard>

\quad [TOUCH] <keyboard>

\item 
\textbf{Task 3:}
I am in [`bathroom’]. The objects I can manipulate are [`dishwasher’, `faucet’, `keyboard’, `television’, `newspaper’, `chair’, `coffe\_maker’, `pillow’, `phone’, `cup’, `check’, `couch’, `freezer’, `desk’, `oven’, `light’, `food\_food’, `plate’, `table’, `bookmark’, `bedroom’, `dining\_room’, `cupboard’, `computer’, `sink’, `bed’, `cat’, `mouse’, `home\_office’, `pot’].

\textbf{Goal:}
Pick up phone

\textbf{Hint:}
First when I hear the ringing sound I will run to my living room and picks
up and I will say hello

\textbf{Plan:}

\quad [RUN] <home\_office>

\quad [WALK] <chair>

\quad [FIND] <chair>

\quad [SIT] <chair>

\quad [FIND] <phone>

\quad [GRAB] <phone>

\item 
\textbf{Task 4:}
\textbf{[QUESTION]}
\end{itemize}
\end{tcolorbox}
\end{tcbraster}

{\small
 \bibliographystyle{ieee}
 \bibliography{egbib}

\begin{thebibliography}{10}\itemsep=-1pt

\bibitem{ahn2022icanisay}
M.~Ahn, A.~Brohan, N.~Brown, Y.~Chebotar, O.~Cortes, B.~David, C.~Finn, C.~Fu, K.~Gopalakrishnan, K.~Hausman, A.~Herzog, D.~Ho, J.~Hsu, J.~Ibarz, B.~Ichter, A.~Irpan, E.~Jang, R.~J. Ruano, K.~Jeffrey, S.~Jesmonth, N.~J. Joshi, R.~Julian, D.~Kalashnikov, Y.~Kuang, K.-H. Lee, S.~Levine, Y.~Lu, L.~Luu, C.~Parada, P.~Pastor, J.~Quiambao, K.~Rao, J.~Rettinghouse, D.~Reyes, P.~Sermanet, N.~Sievers, C.~Tan, A.~Toshev, V.~Vanhoucke, F.~Xia, T.~Xiao, P.~Xu, S.~Xu, M.~Yan, and A.~Zeng.
\newblock Do as i can, not as i say: Grounding language in robotic affordances, 2022.

\bibitem{borgeaud2022improvinglanguagemodelsretrieving}
S.~Borgeaud, A.~Mensch, J.~Hoffmann, T.~Cai, E.~Rutherford, K.~Millican, G.~van~den Driessche, J.-B. Lespiau, B.~Damoc, A.~Clark, D.~de~Las~Casas, A.~Guy, J.~Menick, R.~Ring, T.~Hennigan, S.~Huang, L.~Maggiore, C.~Jones, A.~Cassirer, A.~Brock, M.~Paganini, G.~Irving, O.~Vinyals, S.~Osindero, K.~Simonyan, J.~W. Rae, E.~Elsen, and L.~Sifre.
\newblock Improving language models by retrieving from trillions of tokens, 2022.

\bibitem{brown2020languagemodelsfewshotlearners}
T.~B. Brown, B.~Mann, N.~Ryder, M.~Subbiah, J.~Kaplan, P.~Dhariwal, A.~Neelakantan, P.~Shyam, G.~Sastry, A.~Askell, S.~Agarwal, A.~Herbert-Voss, G.~Krueger, T.~Henighan, R.~Child, A.~Ramesh, D.~M. Ziegler, J.~Wu, C.~Winter, C.~Hesse, M.~Chen, E.~Sigler, M.~Litwin, S.~Gray, B.~Chess, J.~Clark, C.~Berner, S.~McCandlish, A.~Radford, I.~Sutskever, and D.~Amodei.
\newblock Language models are few-shot learners, 2020.

\bibitem{bubeck2023sparksartificialgeneralintelligence}
S.~Bubeck, V.~Chandrasekaran, R.~Eldan, J.~Gehrke, E.~Horvitz, E.~Kamar, P.~Lee, Y.~T. Lee, Y.~Li, S.~Lundberg, H.~Nori, H.~Palangi, M.~T. Ribeiro, and Y.~Zhang.
\newblock Sparks of artificial general intelligence: Early experiments with gpt-4, 2023.

\bibitem{10937907}
W.~Cai, J.~Jiang, F.~Wang, J.~Tang, S.~Kim, and J.~Huang.
\newblock A survey on mixture of experts in large language models.
\newblock {\em IEEE Transactions on Knowledge and Data Engineering}, pages 1--20, 2025.

\bibitem{chen2024agentflandesigningdatamethods}
Z.~Chen, K.~Liu, Q.~Wang, W.~Zhang, J.~Liu, D.~Lin, K.~Chen, and F.~Zhao.
\newblock Agent-flan: Designing data and methods of effective agent tuning for large language models, 2024.

\bibitem{chowdhery2022palmscalinglanguagemodeling}
A.~Chowdhery, S.~Narang, J.~Devlin, M.~Bosma, G.~Mishra, A.~Roberts, P.~Barham, H.~W. Chung, C.~Sutton, S.~Gehrmann, P.~Schuh, K.~Shi, S.~Tsvyashchenko, J.~Maynez, A.~Rao, P.~Barnes, Y.~Tay, N.~Shazeer, V.~Prabhakaran, E.~Reif, N.~Du, B.~Hutchinson, R.~Pope, J.~Bradbury, J.~Austin, M.~Isard, G.~Gur-Ari, P.~Yin, T.~Duke, A.~Levskaya, S.~Ghemawat, S.~Dev, H.~Michalewski, X.~Garcia, V.~Misra, K.~Robinson, L.~Fedus, D.~Zhou, D.~Ippolito, D.~Luan, H.~Lim, B.~Zoph, A.~Spiridonov, R.~Sepassi, D.~Dohan, S.~Agrawal, M.~Omernick, A.~M. Dai, T.~S. Pillai, M.~Pellat, A.~Lewkowycz, E.~Moreira, R.~Child, O.~Polozov, K.~Lee, Z.~Zhou, X.~Wang, B.~Saeta, M.~Diaz, O.~Firat, M.~Catasta, J.~Wei, K.~Meier-Hellstern, D.~Eck, J.~Dean, S.~Petrov, and N.~Fiedel.
\newblock Palm: Scaling language modeling with pathways, 2022.

\bibitem{cobbe2021trainingverifierssolvemath}
K.~Cobbe, V.~Kosaraju, M.~Bavarian, M.~Chen, H.~Jun, L.~Kaiser, M.~Plappert, J.~Tworek, J.~Hilton, R.~Nakano, C.~Hesse, and J.~Schulman.
\newblock Training verifiers to solve math word problems, 2021.

\bibitem{fu2020surveycomplexquestionanswering}
B.~Fu, Y.~Qiu, C.~Tang, Y.~Li, H.~Yu, and J.~Sun.
\newblock A survey on complex question answering over knowledge base: Recent advances and challenges, 2020.

\bibitem{glaese2022improvingalignmentdialogueagents}
A.~Glaese, N.~McAleese, M.~Trębacz, J.~Aslanides, V.~Firoiu, T.~Ewalds, M.~Rauh, L.~Weidinger, M.~Chadwick, P.~Thacker, L.~Campbell-Gillingham, J.~Uesato, P.-S. Huang, R.~Comanescu, F.~Yang, A.~See, S.~Dathathri, R.~Greig, C.~Chen, D.~Fritz, J.~S. Elias, R.~Green, S.~Mokrá, N.~Fernando, B.~Wu, R.~Foley, S.~Young, I.~Gabriel, W.~Isaac, J.~Mellor, D.~Hassabis, K.~Kavukcuoglu, L.~A. Hendricks, and G.~Irving.
\newblock Improving alignment of dialogue agents via targeted human judgements, 2022.

\bibitem{guu2020realmretrievalaugmentedlanguagemodel}
K.~Guu, K.~Lee, Z.~Tung, P.~Pasupat, and M.-W. Chang.
\newblock Realm: Retrieval-augmented language model pre-training, 2020.

\bibitem{hao2024toolkengptaugmentingfrozenlanguage}
S.~Hao, T.~Liu, Z.~Wang, and Z.~Hu.
\newblock Toolkengpt: Augmenting frozen language models with massive tools via tool embeddings, 2024.

\bibitem{hao2023bertnetharvestingknowledgegraphs}
S.~Hao, B.~Tan, K.~Tang, B.~Ni, X.~Shao, H.~Zhang, E.~P. Xing, and Z.~Hu.
\newblock Bertnet: Harvesting knowledge graphs with arbitrary relations from pretrained language models, 2023.

\bibitem{huang-etal-2016-well}
D.~Huang, S.~Shi, C.-Y. Lin, J.~Yin, and W.-Y. Ma.
\newblock How well do computers solve math word problems? large-scale dataset construction and evaluation.
\newblock In K.~Erk and N.~A. Smith, editors, {\em Proceedings of the 54th Annual Meeting of the Association for Computational Linguistics (Volume 1: Long Papers)}, pages 887--896, Berlin, Germany, Aug. 2016. Association for Computational Linguistics.

\bibitem{huang2022languagemodelszeroshotplanners}
W.~Huang, P.~Abbeel, D.~Pathak, and I.~Mordatch.
\newblock Language models as zero-shot planners: Extracting actionable knowledge for embodied agents, 2022.

\bibitem{huang2023groundeddecodingguidingtext}
W.~Huang, F.~Xia, D.~Shah, D.~Driess, A.~Zeng, Y.~Lu, P.~Florence, I.~Mordatch, S.~Levine, K.~Hausman, and B.~Ichter.
\newblock Grounded decoding: Guiding text generation with grounded models for embodied agents, 2023.

\bibitem{huang2022innermonologueembodiedreasoning}
W.~Huang, F.~Xia, T.~Xiao, H.~Chan, J.~Liang, P.~Florence, A.~Zeng, J.~Tompson, I.~Mordatch, Y.~Chebotar, P.~Sermanet, N.~Brown, T.~Jackson, L.~Luu, S.~Levine, K.~Hausman, and B.~Ichter.
\newblock Inner monologue: Embodied reasoning through planning with language models, 2022.

\bibitem{bommarito2022gpttakesbarexam}
M.~B. II and D.~M. Katz.
\newblock Gpt takes the bar exam, 2022.

\bibitem{Ji_2023}
Z.~Ji, N.~Lee, R.~Frieske, T.~Yu, D.~Su, Y.~Xu, E.~Ishii, Y.~J. Bang, A.~Madotto, and P.~Fung.
\newblock Survey of hallucination in natural language generation.
\newblock {\em ACM Computing Surveys}, 55(12):1–38, Mar. 2023.

\bibitem{khattab2023demonstratesearchpredictcomposingretrievallanguage}
O.~Khattab, K.~Santhanam, X.~L. Li, D.~Hall, P.~Liang, C.~Potts, and M.~Zaharia.
\newblock Demonstrate-search-predict: Composing retrieval and language models for knowledge-intensive nlp, 2023.

\bibitem{lewis2021retrievalaugmentedgenerationknowledgeintensivenlp}
P.~Lewis, E.~Perez, A.~Piktus, F.~Petroni, V.~Karpukhin, N.~Goyal, H.~Küttler, M.~Lewis, W.~tau Yih, T.~Rocktäschel, S.~Riedel, and D.~Kiela.
\newblock Retrieval-augmented generation for knowledge-intensive nlp tasks, 2021.

\bibitem{li2023modelscopeagentbuildingcustomizableagent}
C.~Li, H.~Chen, M.~Yan, W.~Shen, H.~Xu, Z.~Wu, Z.~Zhang, W.~Zhou, Y.~Chen, C.~Cheng, H.~Shi, J.~Zhang, F.~Huang, and J.~Zhou.
\newblock Modelscope-agent: Building your customizable agent system with open-source large language models, 2023.

\bibitem{li2023apibankcomprehensivebenchmarktoolaugmented}
M.~Li, Y.~Zhao, B.~Yu, F.~Song, H.~Li, H.~Yu, Z.~Li, F.~Huang, and Y.~Li.
\newblock Api-bank: A comprehensive benchmark for tool-augmented llms, 2023.

\bibitem{liang2023taskmatrixaicompletingtasksconnecting}
Y.~Liang, C.~Wu, T.~Song, W.~Wu, Y.~Xia, Y.~Liu, Y.~Ou, S.~Lu, L.~Ji, S.~Mao, Y.~Wang, L.~Shou, M.~Gong, and N.~Duan.
\newblock Taskmatrix.ai: Completing tasks by connecting foundation models with millions of apis, 2023.

\bibitem{10706809}
Q.~Ma, Z.~Liu, Z.~Zheng, Z.~Huang, S.~Zhu, Z.~Yu, and J.~T. Kwok.
\newblock A survey on time-series pre-trained models.
\newblock {\em IEEE Transactions on Knowledge and Data Engineering}, 36(12):7536--7555, 2024.

\bibitem{mialon2023augmentedlanguagemodelssurvey}
G.~Mialon, R.~Dessì, M.~Lomeli, C.~Nalmpantis, R.~Pasunuru, R.~Raileanu, B.~Rozière, T.~Schick, J.~Dwivedi-Yu, A.~Celikyilmaz, E.~Grave, Y.~LeCun, and T.~Scialom.
\newblock Augmented language models: a survey, 2023.

\bibitem{nakano2022webgptbrowserassistedquestionansweringhuman}
R.~Nakano, J.~Hilton, S.~Balaji, J.~Wu, L.~Ouyang, C.~Kim, C.~Hesse, S.~Jain, V.~Kosaraju, W.~Saunders, X.~Jiang, K.~Cobbe, T.~Eloundou, G.~Krueger, K.~Button, M.~Knight, B.~Chess, and J.~Schulman.
\newblock Webgpt: Browser-assisted question-answering with human feedback, 2022.

\bibitem{10387715}
S.~Pan, L.~Luo, Y.~Wang, C.~Chen, J.~Wang, and X.~Wu.
\newblock Unifying large language models and knowledge graphs: A roadmap.
\newblock {\em IEEE Transactions on Knowledge and Data Engineering}, 36(7):3580--3599, 2024.

\bibitem{parisi2022talmtoolaugmentedlanguage}
A.~Parisi, Y.~Zhao, and N.~Fiedel.
\newblock Talm: Tool augmented language models, 2022.

\bibitem{puig2018virtualhomesimulatinghouseholdactivities}
X.~Puig, K.~Ra, M.~Boben, J.~Li, T.~Wang, S.~Fidler, and A.~Torralba.
\newblock Virtualhome: Simulating household activities via programs, 2018.

\bibitem{qin2023webcpminteractivewebsearch}
Y.~Qin, Z.~Cai, D.~Jin, L.~Yan, S.~Liang, K.~Zhu, Y.~Lin, X.~Han, N.~Ding, H.~Wang, R.~Xie, F.~Qi, Z.~Liu, M.~Sun, and J.~Zhou.
\newblock Webcpm: Interactive web search for chinese long-form question answering, 2023.

\bibitem{qin2024toollearningfoundationmodels}
Y.~Qin, S.~Hu, Y.~Lin, W.~Chen, N.~Ding, G.~Cui, Z.~Zeng, Y.~Huang, C.~Xiao, C.~Han, Y.~R. Fung, Y.~Su, H.~Wang, C.~Qian, R.~Tian, K.~Zhu, S.~Liang, X.~Shen, B.~Xu, Z.~Zhang, Y.~Ye, B.~Li, Z.~Tang, J.~Yi, Y.~Zhu, Z.~Dai, L.~Yan, X.~Cong, Y.~Lu, W.~Zhao, Y.~Huang, J.~Yan, X.~Han, X.~Sun, D.~Li, J.~Phang, C.~Yang, T.~Wu, H.~Ji, Z.~Liu, and M.~Sun.
\newblock Tool learning with foundation models, 2024.

\bibitem{roller2020recipesbuildingopendomainchatbot}
S.~Roller, E.~Dinan, N.~Goyal, D.~Ju, M.~Williamson, Y.~Liu, J.~Xu, M.~Ott, K.~Shuster, E.~M. Smith, Y.-L. Boureau, and J.~Weston.
\newblock Recipes for building an open-domain chatbot, 2020.

\bibitem{rozière2024codellamaopenfoundation}
B.~Rozière, J.~Gehring, F.~Gloeckle, S.~Sootla, I.~Gat, X.~E. Tan, Y.~Adi, J.~Liu, R.~Sauvestre, T.~Remez, J.~Rapin, A.~Kozhevnikov, I.~Evtimov, J.~Bitton, M.~Bhatt, C.~C. Ferrer, A.~Grattafiori, W.~Xiong, A.~Défossez, J.~Copet, F.~Azhar, H.~Touvron, L.~Martin, N.~Usunier, T.~Scialom, and G.~Synnaeve.
\newblock Code llama: Open foundation models for code, 2024.

\bibitem{schick2023toolformerlanguagemodelsteach}
T.~Schick, J.~Dwivedi-Yu, R.~Dessì, R.~Raileanu, M.~Lomeli, L.~Zettlemoyer, N.~Cancedda, and T.~Scialom.
\newblock Toolformer: Language models can teach themselves to use tools, 2023.

\bibitem{shuster2021retrievalaugmentationreduceshallucination}
K.~Shuster, S.~Poff, M.~Chen, D.~Kiela, and J.~Weston.
\newblock Retrieval augmentation reduces hallucination in conversation, 2021.

\bibitem{singh2022progpromptgeneratingsituatedrobot}
I.~Singh, V.~Blukis, A.~Mousavian, A.~Goyal, D.~Xu, J.~Tremblay, D.~Fox, J.~Thomason, and A.~Garg.
\newblock Progprompt: Generating situated robot task plans using large language models, 2022.

\bibitem{talmor2018webknowledgebaseansweringcomplex}
A.~Talmor and J.~Berant.
\newblock The web as a knowledge-base for answering complex questions, 2018.

\bibitem{tang2023toolalpacageneralizedtoollearning}
Q.~Tang, Z.~Deng, H.~Lin, X.~Han, Q.~Liang, B.~Cao, and L.~Sun.
\newblock Toolalpaca: Generalized tool learning for language models with 3000 simulated cases, 2023.

\bibitem{thoppilan2022lamdalanguagemodelsdialog}
R.~Thoppilan, D.~D. Freitas, J.~Hall, N.~Shazeer, A.~Kulshreshtha, H.-T. Cheng, A.~Jin, T.~Bos, L.~Baker, Y.~Du, Y.~Li, H.~Lee, H.~S. Zheng, A.~Ghafouri, M.~Menegali, Y.~Huang, M.~Krikun, D.~Lepikhin, J.~Qin, D.~Chen, Y.~Xu, Z.~Chen, A.~Roberts, M.~Bosma, V.~Zhao, Y.~Zhou, C.-C. Chang, I.~Krivokon, W.~Rusch, M.~Pickett, P.~Srinivasan, L.~Man, K.~Meier-Hellstern, M.~R. Morris, T.~Doshi, R.~D. Santos, T.~Duke, J.~Soraker, B.~Zevenbergen, V.~Prabhakaran, M.~Diaz, B.~Hutchinson, K.~Olson, A.~Molina, E.~Hoffman-John, J.~Lee, L.~Aroyo, R.~Rajakumar, A.~Butryna, M.~Lamm, V.~Kuzmina, J.~Fenton, A.~Cohen, R.~Bernstein, R.~Kurzweil, B.~Aguera-Arcas, C.~Cui, M.~Croak, E.~Chi, and Q.~Le.
\newblock Lamda: Language models for dialog applications, 2022.

\bibitem{touvron2023llamaopenefficientfoundation}
H.~Touvron, T.~Lavril, G.~Izacard, X.~Martinet, M.-A. Lachaux, T.~Lacroix, B.~Rozière, N.~Goyal, E.~Hambro, F.~Azhar, A.~Rodriguez, A.~Joulin, E.~Grave, and G.~Lample.
\newblock Llama: Open and efficient foundation language models, 2023.

\bibitem{touvron2023llama2openfoundation}
H.~Touvron, L.~Martin, K.~Stone, P.~Albert, A.~Almahairi, Y.~Babaei, N.~Bashlykov, S.~Batra, P.~Bhargava, S.~Bhosale, D.~Bikel, L.~Blecher, C.~C. Ferrer, M.~Chen, G.~Cucurull, D.~Esiobu, J.~Fernandes, J.~Fu, W.~Fu, B.~Fuller, C.~Gao, V.~Goswami, N.~Goyal, A.~Hartshorn, S.~Hosseini, R.~Hou, H.~Inan, M.~Kardas, V.~Kerkez, M.~Khabsa, I.~Kloumann, A.~Korenev, P.~S. Koura, M.-A. Lachaux, T.~Lavril, J.~Lee, D.~Liskovich, Y.~Lu, Y.~Mao, X.~Martinet, T.~Mihaylov, P.~Mishra, I.~Molybog, Y.~Nie, A.~Poulton, J.~Reizenstein, R.~Rungta, K.~Saladi, A.~Schelten, R.~Silva, E.~M. Smith, R.~Subramanian, X.~E. Tan, B.~Tang, R.~Taylor, A.~Williams, J.~X. Kuan, P.~Xu, Z.~Yan, I.~Zarov, Y.~Zhang, A.~Fan, M.~Kambadur, S.~Narang, A.~Rodriguez, R.~Stojnic, S.~Edunov, and T.~Scialom.
\newblock Llama 2: Open foundation and fine-tuned chat models, 2023.

\bibitem{9782500}
J.~Wang, C.~Lan, C.~Liu, Y.~Ouyang, T.~Qin, W.~Lu, Y.~Chen, W.~Zeng, and P.~S. Yu.
\newblock Generalizing to unseen domains: A survey on domain generalization.
\newblock {\em IEEE Transactions on Knowledge and Data Engineering}, 35(8):8052--8072, 2023.

\bibitem{wei2023chainofthoughtpromptingelicitsreasoning}
J.~Wei, X.~Wang, D.~Schuurmans, M.~Bosma, B.~Ichter, F.~Xia, E.~Chi, Q.~Le, and D.~Zhou.
\newblock Chain-of-thought prompting elicits reasoning in large language models, 2023.

\bibitem{xiang2023languagemodelsmeetworld}
J.~Xiang, T.~Tao, Y.~Gu, T.~Shu, Z.~Wang, Z.~Yang, and Z.~Hu.
\newblock Language models meet world models: Embodied experiences enhance language models, 2023.

\bibitem{yang2018hotpotqadatasetdiverseexplainable}
Z.~Yang, P.~Qi, S.~Zhang, Y.~Bengio, W.~W. Cohen, R.~Salakhutdinov, and C.~D. Manning.
\newblock Hotpotqa: A dataset for diverse, explainable multi-hop question answering, 2018.

\bibitem{yao2020calmexplorelanguagemodels}
S.~Yao, R.~Rao, M.~Hausknecht, and K.~Narasimhan.
\newblock Keep calm and explore: Language models for action generation in text-based games, 2020.

\bibitem{yao2023reactsynergizingreasoningacting}
S.~Yao, J.~Zhao, D.~Yu, N.~Du, I.~Shafran, K.~Narasimhan, and Y.~Cao.
\newblock React: Synergizing reasoning and acting in language models, 2023.

\bibitem{yao2024retroformerretrospectivelargelanguage}
W.~Yao, S.~Heinecke, J.~C. Niebles, Z.~Liu, Y.~Feng, L.~Xue, R.~Murthy, Z.~Chen, J.~Zhang, D.~Arpit, R.~Xu, P.~Mui, H.~Wang, C.~Xiong, and S.~Savarese.
\newblock Retroformer: Retrospective large language agents with policy gradient optimization, 2024.

\bibitem{ye2024tooleyesfinegrainedevaluationtool}
J.~Ye, G.~Li, S.~Gao, C.~Huang, Y.~Wu, S.~Li, X.~Fan, S.~Dou, T.~Ji, Q.~Zhang, T.~Gui, and X.~Huang.
\newblock Tooleyes: Fine-grained evaluation for tool learning capabilities of large language models in real-world scenarios, 2024.

\bibitem{Yu_2022}
W.~Yu, C.~Zhu, Z.~Li, Z.~Hu, Q.~Wang, H.~Ji, and M.~Jiang.
\newblock A survey of knowledge-enhanced text generation.
\newblock {\em ACM Computing Surveys}, 54(11s):1–38, Jan. 2022.

\bibitem{zeng2022socraticmodelscomposingzeroshot}
A.~Zeng, M.~Attarian, B.~Ichter, K.~Choromanski, A.~Wong, S.~Welker, F.~Tombari, A.~Purohit, M.~Ryoo, V.~Sindhwani, J.~Lee, V.~Vanhoucke, and P.~Florence.
\newblock Socratic models: Composing zero-shot multimodal reasoning with language, 2022.

\bibitem{zha2023alignscoreevaluatingfactualconsistency}
Y.~Zha, Y.~Yang, R.~Li, and Z.~Hu.
\newblock Alignscore: Evaluating factual consistency with a unified alignment function, 2023.

\bibitem{zha2023textalignmentefficientunified}
Y.~Zha, Y.~Yang, R.~Li, and Z.~Hu.
\newblock Text alignment is an efficient unified model for massive nlp tasks, 2023.

\bibitem{10506571}
Z.~Zhao, W.~Fan, J.~Li, Y.~Liu, X.~Mei, Y.~Wang, Z.~Wen, F.~Wang, X.~Zhao, J.~Tang, and Q.~Li.
\newblock Recommender systems in the era of large language models (llms).
\newblock {\em IEEE Transactions on Knowledge and Data Engineering}, 36(11):6889--6907, 2024.

\bibitem{zhong2023memorybankenhancinglargelanguage}
W.~Zhong, L.~Guo, Q.~Gao, H.~Ye, and Y.~Wang.
\newblock Memorybank: Enhancing large language models with long-term memory, 2023.

\bibitem{zong2024triadframeworkleveragingmultirole}
C.~Zong, Y.~Yan, W.~Lu, J.~Shao, E.~Huang, H.~Chang, and Y.~Zhuang.
\newblock Triad: A framework leveraging a multi-role llm-based agent to solve knowledge base question answering, 2024.

\end{thebibliography}
}
\end{document}